\documentclass[10pt,twocolumn,letterpaper]{article}

\usepackage{iccv}
\usepackage{times}
\usepackage{epsfig}
\usepackage{graphicx}
\usepackage{amsmath}
\usepackage{amssymb}

\usepackage{booktabs}
\usepackage{multirow}
\usepackage{makecell}
\usepackage{nicefrac}
\usepackage{diagbox}
\usepackage{enumitem}
\usepackage{authblk}

\usepackage{caption}
\usepackage[table]{xcolor}         
\definecolor{codegreen}{rgb}{0,0.6,0}
\definecolor{codegray}{rgb}{0.5,0.5,0.5}
\definecolor{codepurple}{rgb}{0.58,0,0.82}
\definecolor{backcolour}{rgb}{0.95,0.95,0.92}
\definecolor{Blue}{rgb}{0,0,1}
\definecolor{Green}{rgb}{0,1,0}

\usepackage{soul}

\usepackage{amsthm}

\newtheorem{thm}{Theorem}
\usepackage{enumitem}

\usepackage[pagebackref=true,breaklinks=true,letterpaper=true,colorlinks,bookmarks=false]{hyperref}

\iccvfinalcopy 

\def\iccvPaperID{6910} 

\ificcvfinal\fi

\begin{document}

\title{Householder Projector for Unsupervised Latent Semantics Discovery}

\author[1]{{Yue Song}}
\author[1]{{Jichao Zhang}}
\author[1]{{Nicu Sebe}}
\author[2]{{Wei Wang}}
\affil[1]{Department of Information Engineering and Computer Science, University of Trento, Italy}
\affil[2]{Beijing Jiaotong University, China}
\affil[ ]{\texttt{yue.song@unitn.it}}

\maketitle
\ificcvfinal\thispagestyle{empty}\fi

\begin{abstract}
   Generative Adversarial Networks (GANs), especially the recent style-based generators (StyleGANs), have versatile semantics in the structured latent space. Latent semantics discovery methods emerge to move around the latent code such that only one factor varies during the traversal. Recently, an unsupervised method proposed a promising direction to directly use the eigenvectors of the projection matrix that maps latent codes to features as the interpretable directions. However, one overlooked fact is that the projection matrix is non-orthogonal and the number of eigenvectors is too large. The non-orthogonality would entangle semantic attributes in the top few eigenvectors, and the large dimensionality might result in meaningless variations among the directions even if the matrix is orthogonal. To avoid these issues, we propose Householder Projector, a flexible and general low-rank orthogonal matrix representation based on Householder transformations, to parameterize the projection matrix. The orthogonality guarantees that the eigenvectors correspond to disentangled interpretable semantics, while the low-rank property encourages that each identified direction has meaningful variations. We integrate our projector into pre-trained StyleGAN2/StyleGAN3 and evaluate the models on several benchmarks. Within only $1\%$ of the original training steps for fine-tuning, our projector helps StyleGANs to discover more disentangled and precise semantic attributes without sacrificing image fidelity. Code is publicly available via \url{https://github.com/KingJamesSong/HouseholderGAN}.\let\thefootnote\relax\footnotetext{The first two authors contribute equally; Wei Wang is the corresponding author.}
\end{abstract}

\section{Introduction}

\begin{figure}[htbp]
    \centering
    \includegraphics[width=0.99\linewidth]{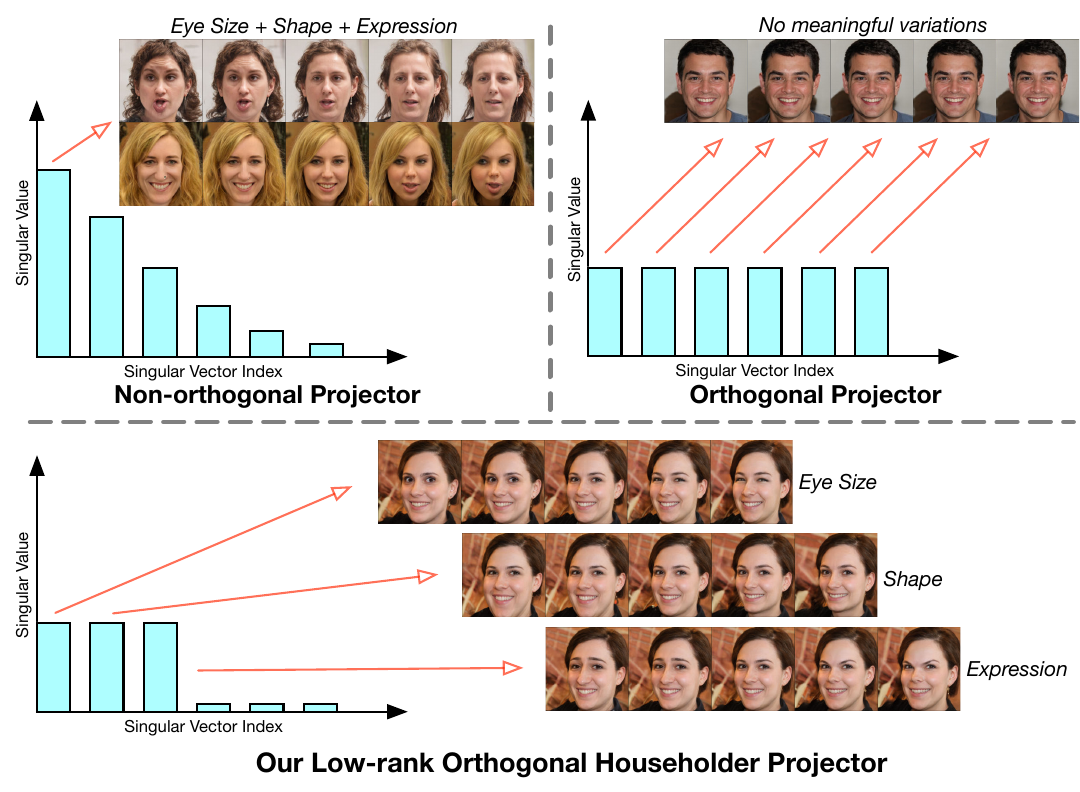}
    \caption{Motivation of our proposed Householder Projector. \textbf{\textit{Here ``Projector" denotes the projection matrix that maps latent codes to features, \emph{i.e.,} the modulation weight of StyleGANs.}} (\textit{Top Left}) The singular value imbalance of the non-orthogonal projector would entangle multiple semantics in the top interpretable directions. (\textit{Top Right}) Due to the large dimensionality of the projector, directly enforcing vanilla orthogonality would spread the data variations among all the eigenvectors, leading to imperceptible and meaningless traversal. \textbf{(\textit{Bottom})} Our Householder Projector equips the projection matrix with low-rank orthogonal properties, which simultaneously disentangles semantics into multiple equally-important eigenvectors and guarantees that each direction could correspond to semantically-meaningful variations. }
    \label{fig:motivation}
    \vspace{-0.2cm}
\end{figure}

Generative Adversarial Networks (GANs)~\cite{goodfellow2014generative} and the recent style-based generative models (StyleGANs)~\cite{karras2019style,karras2020analyzing,karras2021alias} in particular have become the leading paradigm of generative modeling in the vision domain. The latent spaces of StyleGANs are known to embed rich and hierarchical semantics~\cite{goetschalckx2019ganalyze,jahanian2020steerability}; moving the latent code in certain directions could trigger meaningful variations in the output images. Therefore, latent semantics discovery methods emerge to identify such interpretable directions that each variation factor is disentangled and the generation process can be precisely controled~\cite{goetschalckx2019ganalyze,shen2020interpreting,plumerault2020controlling,yang2021semantic,shen2021closed,abdal2021styleflow,wu2021stylespace,oldfield2021tensor,chen2022exploring}. 



Among the recent explorations of unsupervised interpretable semantics discovery methods~\cite{voynov2020unsupervised,shen2021closed,wei2021orthogonal,peebles2020hessian}, SeFa~\cite{shen2021closed} pointed out a promising direction to discover semantically meaningful concepts by computing the eigenvector of the projector. Here we refer to the projection matrix that maps latent codes to features as the projector. The key observation is that using the eigenvectors of the projector for latent perturbation would maximize the data variations. Such identified eigenvectors/directions would correspond to meaningful semantic concepts. However, as shown in Fig.~\ref{fig:motivation} top left, this is likely to cause semantics entangled in the top few eigenvectors. This phenomenon stems from the fact that the variation caused by an eigenvector is actually determined by the associated eigenvalue. Due to the imbalanced eigenvalue distribution, the discovered directions are not equally-important, and the top few ones would simultaneously manipulate multiple attributes. This eigenvalue discrepancy can be mitigated by enforcing orthogonality constraint to the projector. Nonetheless, since standard orthogonal matrices have as many equally-important eigenvectors as the dimensionality, there would not be enough semantics to mine in practice when the method scales to large models such as StyleGANs whose projector dimension is $1024$ or $512$. Consequently, as an accompanying limitation, the data variations would be split among all the eigenvectors and none of them could produce meaningful output changes (see Fig.~\ref{fig:motivation} top right).

To resolve the above issues, we propose Householder Projector, a low-rank orthogonal matrix representation based on Householder transformations, to parameterize the projection matrix that maps latent codes to features. The projector is first decomposed to its SVD form ($\mathbf{U}\mathbf{S}\mathbf{V}^{T}$). Next, the orthogonal singular vectors $\mathbf{U}$ and $\mathbf{V}$ are represented by a series of Householder reflectors, respectively. Thanks to the normalization of Householder reflections, the orthogonality is also preserved during backpropagation. For the singular value $\mathbf{S}$, we explicitly set it as a low-rank identity matrix (\emph{i.e.,} $\mathbf{S}{=}{\rm diag}(1,\dots,1,0,\dots,0)$) whose rank defines exactly the number of semantic concepts. As shown in Fig.~\ref{fig:motivation} bottom, the low-rank property guarantees that the identified directions would cause meaningful variations, while the orthogonality encourages that each semantic attribute is disentangled from the others. Moreover, a proper initialization scheme is proposed to leverage the statistics of pre-trained weights, and an acceleration technique is applied to speed up the computation. We also propose a metric dedicated to measuring the smoothness of latent space to interpretable directions based perturbations. Our Householder Projector is integrated into pre-trained StyleGANs~\cite{karras2020analyzing,karras2021alias} at multiple different layers to mine the diverse and hierarchical semantics. Since our projector inevitably changes the pre-trained parameters, the modified models incorporated with our projector are fine-tuned for limited steps to maintain the original image fidelity. Both quantitative and qualitative results on several widely used benchmarks~\cite{kazemi2014one,yu2015lsun,choi2020stargan,karras2020training,fu2022styleganhuman} show that \textbf{within marginal fine-tuning steps ($1\%$ of the training steps), our Householder Projector improves the latent semantics discovery of StyleGANs to have more precise attribute control while not impairing the quality of generated images.} 

The key results and main contributions are as follows:
\begin{itemize}[leftmargin=*]
\vspace{-0.3cm}
    \item We propose Householder Projector, a flexible and general low-rank orthogonal matrix representation based on Householder transformations, to parameterize the projector of generative models for latent semantics discovery. 
    \item Our Householder Projector can be easily integrated into pre-trained GAN models. With the acceleration technique, it can be efficiently fine-tuned for very limited additional steps, which paves the way for applying latent semantics discovery and orthogonality techniques to large-scale  generative models such as StyleGANs. 
\vspace{-0.2cm}
    \item Extensive experiments on two popular backbones (StyleGAN2~\cite{karras2020analyzing} and StyleGAN3~\cite{karras2021alias}) and six benchmarks (FFHQ~\cite{kazemi2014one}, LSUN Church and Cat~\cite{yu2015lsun}, MetFaces~\cite{karras2020training}, AFHQ~\cite{choi2020stargan}, and SHHQ~\cite{fu2022styleganhuman}) demonstrate that within marginal extra fine-tuning steps ($1\%$ of the original training steps), our method could both achieve precise attribute control and preserve the original image quality. 
\end{itemize}





\section{Related Work}

\noindent\textbf{Generative Adversarial Networks.} In the past few years, GAN-based generative models~\cite{goodfellow2014generative} have achieved remarkable progress in high-fidelity image synthesis~\cite{radford2015unsupervised,karras2018progressive,brock2019large,karras2019style,karras2020analyzing,karras2021alias,karras2020training,or2022stylesdf,Chan2022,gu2021stylenerf,skorokhodov2022stylegan,sauer2022stylegan}. The generation process usually takes the following procedure: a randomly-sampled latent code is fed into the generator through a projection step, and then the generator outputs realistic-looking images. Recently, the style-based generators~\cite{karras2019style,karras2020analyzing,karras2021alias} that gradually absorb layer-wise latent style codes are becoming the \emph{de facto} GAN backbones. StyleGAN2~\cite{karras2020analyzing} improves the original StyleGAN~\cite{karras2019style} by redesigning generator normalization and training techniques, and StyleGAN3~\cite{karras2021alias} further explores some equivariance properties. We use the popular StyleGAN2 and StyleGAN3 as our backbones in this paper.


%
\noindent\textbf{Latent Semantics Discovery.} Recently, lots of methods explore disentangling the latent space to achieve image editing by moving the latent code in the identified interpretable directions~\cite{chen2016infogan,bau2019gan,jahanian2020steerability,shen2020interpreting,harkonen2020ganspace,voynov2020unsupervised,zhu2020learning,shen2021closed,he2021eigengan,peebles2020hessian,spingarn2021gan,wei2021orthogonal,tzelepis2021warpedganspace,kim2021exploiting,tzelepis2021warpedganspace,kwon2021diagonal,xu2022transeditor,kappiyath2022self,ren2022learning,choi2022not,song2022orthogonal,song2023latent}. Supervised methods rely on human annotations (\emph{i.e.}, segmentation masks, attribute categories, 3D priors, and text descriptions) to define the semantic labels~\cite{goetschalckx2019ganalyze,shen2020interpreting,yang2021semantic,jahanian2020steerability,plumerault2020controlling,li2021surrogate,chen2022exploring,deng2020disentangled,shi2022semanticstylegan,patashnik2021styleclip,ling2021editgan,yang2021discovering,xu2021linear,karmali2022hierarchical}. Here we mainly highlight some relevant unsupervised approaches that are free of annotations. Voynov~\emph{et al.}~\cite{voynov2020unsupervised} proposed to jointly learn a set of directions and a classifier such that the interpretable directions can be recognized by the classifier. More recently, Peebles~\emph{et al.}~\cite{peebles2020hessian} and Wei~\emph{et al.}~\cite{wei2021orthogonal} proposed to add orthogonal Hessian and Jacobian regularization to encourage disentangled representations, respectively. Song~\emph{et al.}~\cite{song2023latent} proposed to use wave equations to model the spatiotemporal dynamic non-linear latent traversals in generative models.
SeFa~\cite{shen2021closed} showed that the eigenvectors of the projector after the latent code could maximize the data variations and proposed to directly use them as the interpretable directions. However, due to the imbalanced eigenvalues, the image attributes would be entangled in the top few eigenvectors. Our proposed Householder Projector solved this issue by parameterizing the projector to a low-rank but orthogonal matrix. Notice that the used orthogonality technique in~\cite{voynov2020unsupervised} is different from ours. In~\cite{voynov2020unsupervised}, the authors use matrix exponential $\exp(\mathbf{A}{-}\mathbf{A}^{T})$ to generate skew orthogonal matrices where the skew-symmetry could limit the representation power. Also, the technique cannot parameterize given matrices and cannot explicitly control the rank. Our Householder representation is more general and flexible, allowing for controllable rank and parameterization of given matrices. Compared with the \emph{soft} orthogonality regularization used in~\cite{peebles2020hessian,wei2021orthogonal,he2021eigengan}, the \emph{hard} orthogonality of our method helps the model to learn more uncorrelated attributes within less fine-tuning steps.  

In contrast to global editing approaches discussed above, some methods can perform local image editing in a \emph{post hoc} way: they first define or learn a segment of regions of interests, and then manipulate the masked intermediate features for local editing~\cite{bau2020semantic,collins2020editing,zhu2021low,zhu2022region,oldfield2022panda}. Empowered by the precise control of attributes, our method can also achieve competitive performance in local image editing (see Sec.~\ref{sec:quali}). 









\section{Householder Projector}
This section starts with the preliminary introduction of the closed-form latent semantics discovery. Next, we analyze its inherent limitation on entangled semantics and then illustrate our proposed Householder Projector in detail. 

\begin{figure*}[htbp]
    \centering
    \includegraphics[width=0.99\linewidth]{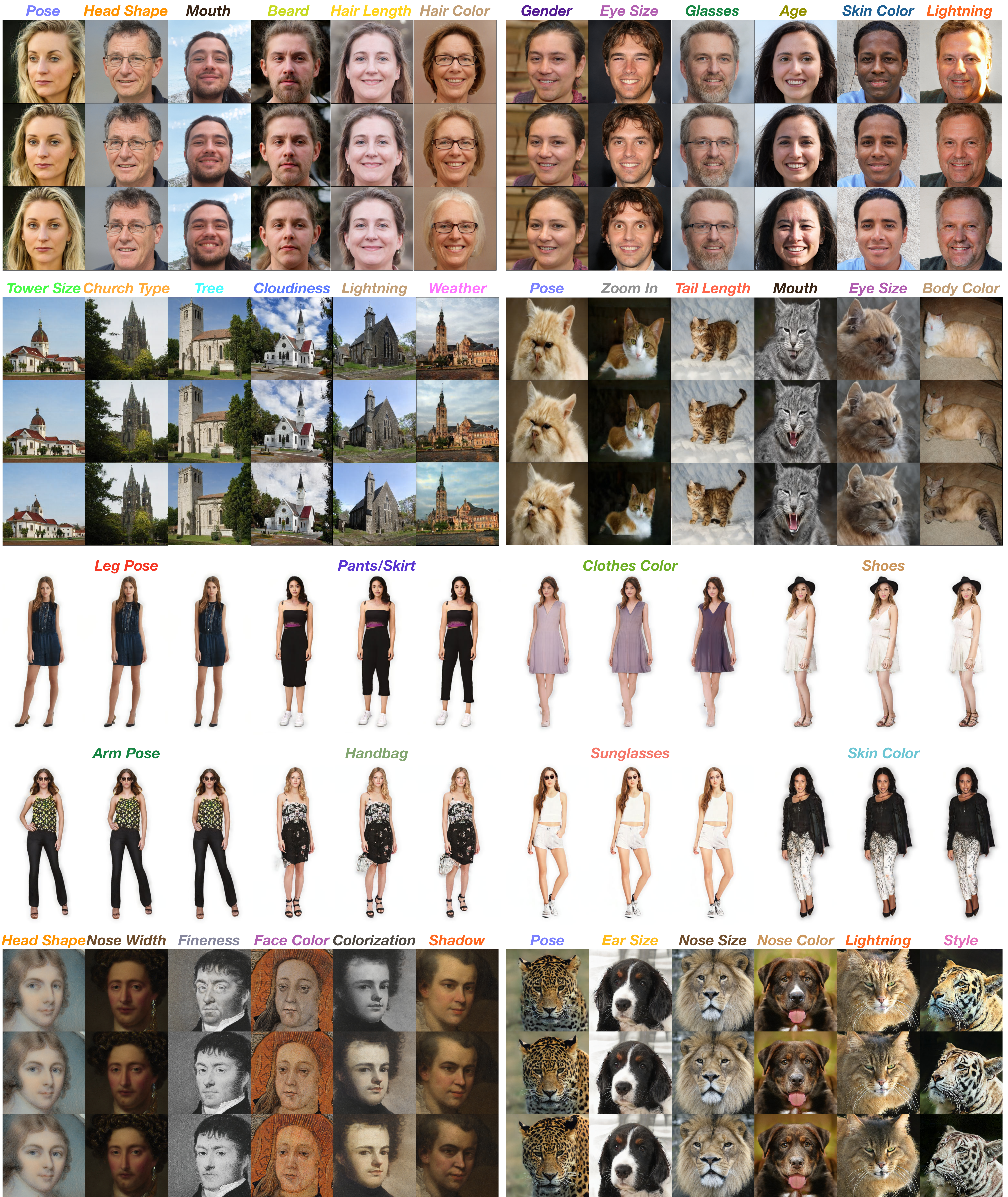}
    \caption{Gallery of some semantic attributes discovered by our Householder Projector across all used datasets (FFHQ~\cite{kazemi2014one} in the top row, LSUN Church~\cite{yu2015lsun} and LSUN Cat~\cite{yu2015lsun} in the $2_{nd}$ row, SHHQ~\cite{fu2022styleganhuman} in the $3_{rd}$ row, and MetFaces~\cite{karras2020training} and AFHQ~\cite{choi2020stargan} in the bottom row). These semantic attributes are sorted from low-level layers (left) to high-level layers (right).} 
    \label{fig:diversity}
\end{figure*}

\subsection{Preliminary: Closed-form Latent Discovery}

Previous latent semantics discovery approaches~\cite{goetschalckx2019ganalyze,voynov2020unsupervised,harkonen2020ganspace,shen2020interpreting,shen2021closed} consider the GAN manipulation as $\texttt{edit}(G(\mathbf{z})){=}G(\mathbf{z}+\alpha\mathbf{n})$ where $G(\cdot)$ represents the generator, $\mathbf{z}{\in}\mathbb{R}^{d}$ denotes the latent code of dimension $d$, $\mathbf{n}{\in}\mathbb{R}^{d}$ is the identified semantically meaningful direction, and $\alpha$ represents the perturbation strength. If one views the GAN as a multi-step projection function, the first projection can be expressed as:
\begin{equation}
\setlength{\abovedisplayskip}{3pt}
\setlength{\belowdisplayskip}{3pt}
    G_{1}(\mathbf{z}+\alpha\mathbf{n}) = \mathbf{A}\mathbf{z} +\mathbf{b} + \alpha\mathbf{A}\mathbf{n},
    \label{eq:edit_affine1}
\end{equation}
where $\mathbf{A}$ and $\mathbf{b}$ denote the weight and bias of the projection step (\emph{e.g.,} convolution or linear transform), respectively. As can be seen from eq.~\eqref{eq:edit_affine1}, the resultant manipulation depends on the term $\alpha\mathbf{A}\mathbf{n}$. Intuitively, an interpretable direction $\mathbf{n}$ should cause large variations of $G_{1}(\mathbf{z}+\alpha\mathbf{n})$, which is equivalent to maximizing the impact of $\alpha\mathbf{A}\mathbf{n}$. Motivated by this observation, SeFa~\cite{shen2021closed} proposed to consider the formulation as the following constrained optimization problem:
\begin{equation}
\setlength{\abovedisplayskip}{3pt}
\setlength{\belowdisplayskip}{3pt}
    \mathbf{n}^{\star} = \arg\max ||\mathbf{A}\mathbf{n}||_{2}^{2}\  s.t. \  \mathbf{n}^{T}\mathbf{n}=1,
\end{equation}
where the constraint $\mathbf{n}^{T}\mathbf{n}{=}1$ is set to satisfy vector orthogonality, and $||\cdot||$ denotes the $l_{2}$ norm. Introducing a Lagrange multiplier $\lambda$ leads to $2\mathbf{A}^{T}\mathbf{A}\mathbf{n}{-}2\lambda\mathbf{n}{=}0$. The closed-form solutions all correspond to the eigenvectors of $\mathbf{A}^{T}\mathbf{A}$. This presents a promising approach to identify the semantics by exploiting the projector $\mathbf{A}$ that projects latent codes. However, one fact overlooked by~\cite{shen2021closed} is that the eigenvectors would cause different extents of variations due to the discrepancy of associated eigenvalues. Supposing that $\mathbf{n}$ is an eigenvector of $\mathbf{A}^{T}\mathbf{A}$, then we would have $||\mathbf{A}\mathbf{n}||_{2}^{2}{=}\sigma^2$ where $\sigma$ is the corresponding singular value of $\mathbf{A}$. For non-orthogonal matrices, the singular values are exponentially decreasing, \emph{i.e.,} $\sigma_{1}{\geq}\sigma_{2}{\geq}\dots{\geq}\sigma_{d}$. This would cause most variations captured by the first few interpretable directions. The imbalance is thus likely to make semantic attributes entangled in the top eigenvectors (see Fig.~\ref{fig:motivation} top left).



\subsection{Householder Low-rank Orthogonal Projector}

The eigenvalue discrepancy can be eliminated by enforcing \textit{strict} orthogonality. Orthogonal matrices have the property of identical eigenvalues, which assigns equal importance to each eigenvector. The low-rank constraint could further limit the number of semantics to mine. Therefore, we propose to use Householder representation, a flexible and general framework to parameterize matrices, to endow the projection matrix with low-rank orthogonality. 

\noindent\textbf{Householder Parameterization.} Householder representations can parameterize any matrices by using a series of Householder reflectors to represent the orthogonal singular vectors of its Singular Value Decomposition (SVD) form. In the field of deep learning, it has been used to parameterize the transition matrix and to stabilize gradients of recurrent neural networks~\cite{mhammedi2017efficient,zhang2018stabilizing,mathiasen2020if}. The key to the orthogonality representation relies on the following theorem:

\begin{thm}[Householder representation~\cite{householder1958unitary,lehoucq1996computation}]
Given any square orthogonal matrix $\mathbf{M}\in\mathbb{R}^{m{\times}m}$, it can be represented by the product of Householder matrices $\mathbf{M}=\mathbf{H}_{1}\mathbf{H}_{2}\dots\mathbf{H}_{m}$ where each Householder matrix is parameterized by a vector as $\mathbf{H}_{i} = \mathbf{I} - 2 \frac{\mathbf{h}_{i}\mathbf{h}_{i}^{T}}{||\mathbf{h}_{i}||_{2}^{2}}$. 
\end{thm}

Let $\mathbf{U}\mathbf{S}\mathbf{V}^{T}$ denote the SVD of the projector $\mathbf{A}$ where $\mathbf{S}$ denotes the diagonal singular value, and $\mathbf{U}$ and $\mathbf{V}$ are left and right orthogonal singular vectors. We use the accumulation of Householder reflectors (\emph{i.e.,} $\prod_{i=0}^{w}\Big(\mathbf{I}{-}2 \frac{\mathbf{u}_{i}\mathbf{u}_{i}^{T}}{||\mathbf{u}_{i}||_{2}^{2}}\Big)$ and  $\prod_{i=0}^{h}\Big(\mathbf{I}{-}2 \frac{\mathbf{v}_{i}\mathbf{v}_{i}^{T}}{||\mathbf{v}_{i}||_{2}^{2}}\Big)$ where $w,h$ denotes the width and height of $\mathbf{A}$) to parameterize $\mathbf{U}$ and $\mathbf{V}$, respectively. Notice that only $\mathbf{u}_{i}$ and $\mathbf{v}_{i}$ are the actual learnable parameters. For $\mathbf{S}$, we explicitly set it to a diagonal matrix and keep the weight fixed. The projector is thus represented by our Householder parameterizations.

\noindent\textbf{Low-rank Constraint.} To achieve the property of disentangled attributes, one straightforward approach is to parameterize $\mathbf{A}$ as an orthogonal matrix, \emph{i.e.,} to set $\mathbf{S}$ to an identity matrix $\mathbf{I}$ where $\mathbf{I}_{i,j}{=}1$ for $i{=}j$ and $\mathbf{I}_{i,j}{=}0$ otherwise. This would lead to equally-important semantic attributes whose number is exactly the projector dimension. However, for large generative models such as StyleGANs, the projector dimension is typically very large (\emph{e.g.,} $512$ or $1024$). It is not likely to have enough attributes to edit in practice. Setting $\mathbf{S}$ to a full-rank diagonal matrix would spread data variations among all the eigenvectors, resulting in trivial and imperceptible traversal (see Fig.~\ref{fig:motivation} top right). To avoid this issue, we propose to define $\mathbf{S}$ as a low-rank identity matrix: 
\begin{equation}
\setlength{\abovedisplayskip}{3pt}
\setlength{\belowdisplayskip}{3pt}
    \mathbf{S}={\rm diag}(\underbrace{1,\dots,1}_{N},0,\dots,0),
\end{equation}
where $N$ defines the rank and also the number of semantic attributes to mine. By restricting the rank of the projector, we explicitly limit the number of interpretable directions. As shown in Fig.~\ref{fig:motivation} bottom, this would benefit the latent traversal for more meaningful output variations.

\noindent\textbf{Orthogonality Preservation.} One advantage of our Householder representation is that the orthogonality can be kept during backpropagation. Since we have the vector normalization $\frac{\mathbf{h}_{i}\mathbf{h}_{i}^{T}}{||\mathbf{h}_{i}||_{2}^{2}}$, the impact of any gradient descent step $\mathbf{h}_{i}{-}\eta\triangledown\mathbf{h}_{i}$ on the orthogonality would be cancelled, \emph{i.e.,} the updated vector remains orthogonal after normalization. 

\begin{figure}[t]
    \centering
    \includegraphics[width=0.99\linewidth]{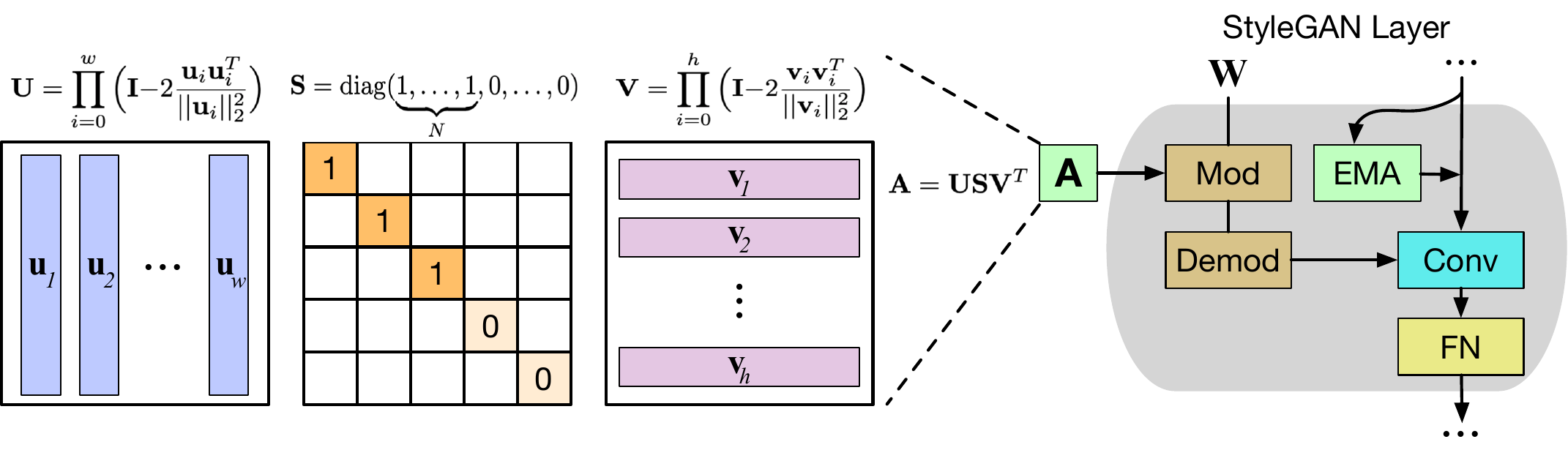}
    \caption{Illustration on how our Householder Projector represents the modulation weight $\mathbf{A}$ of StyleGANs. Here ``Demod", ``EMA", and ``FN'' denote Demodulation, Exponential Moving Average, and Filtered Non-linearities, respectively. The projector is parameterized by its SVD form where $\mathbf{U}$ and $\mathbf{V}$ are represented by the accumulation of Householder reflectors, and $\mathbf{S}$ is set to a low-rank identity matrix. Our projector is applied at multiple different layers of StyleGANs to explore the diverse and hierarchical semantics. The actual learnable parameters are $\mathbf{u}_{i}$ and $\mathbf{v}_{i}$.}
    \label{fig:model}
\end{figure}

\begin{figure}[h]
    \vspace{-0.2cm}
    \centering
    \includegraphics[width=0.99\linewidth]{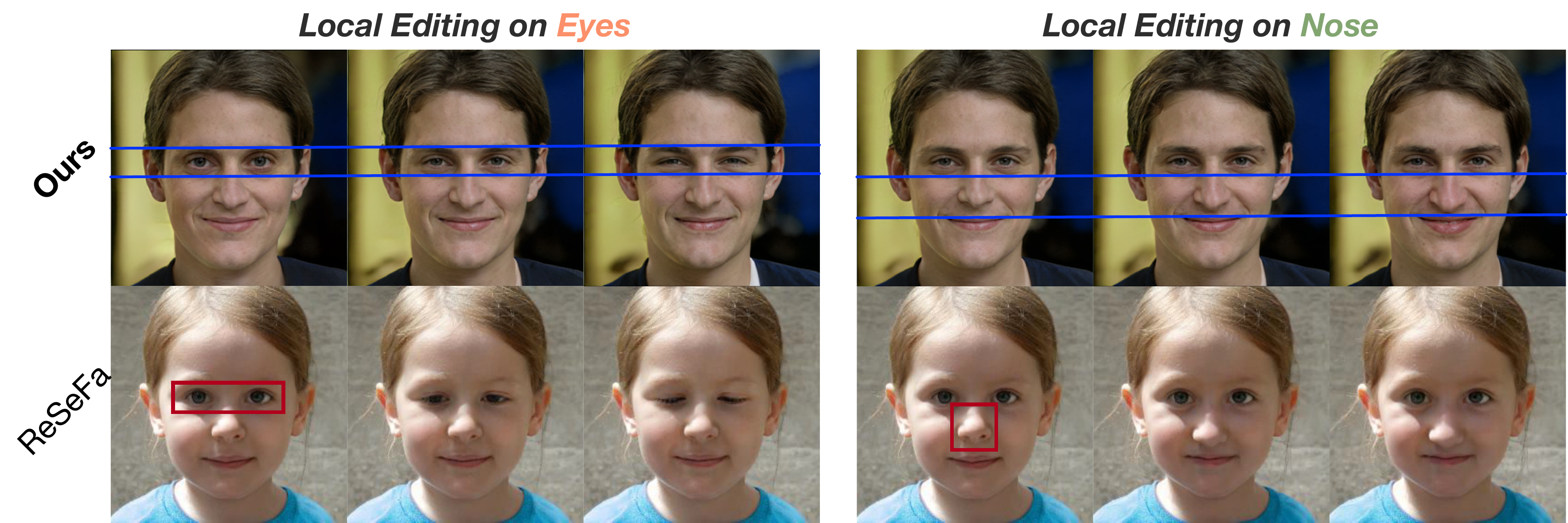}
    \caption{Comparison with ReSeFa~\cite{zhu2022region}. The blue lines indicate the specific regions changed by our method, and the red box indicates the region of interest that is needed as input to ReSeFa~\cite{zhu2022region}.}
    \label{fig:local}
    \vspace{-0.2cm}
\end{figure}

\noindent\textbf{Initialization.} When our method is applied to pre-trained models, the well-trained network weights could be leveraged to initialize our Householder Projector. To this end, we propose to use the nearest-orthogonal mapping~\cite{song2022improving} to project the weight matrix into its orthogonal form that has the nearest distance in the Frobenius norm (\emph{i.e.,} $\min||\mathbf{R}-\mathbf{A}||_{\rm F}$ where $\mathbf{R}^{T}\mathbf{R}=\mathbf{I}$). The closed-form solution is given by $\mathbf{U}\mathbf{V}^{T}$ where $\mathbf{U}\mathbf{S}\mathbf{V}^{T}$ is the SVD of the original weight matrix $\mathbf{A}$. Next, we decompose $\mathbf{U}$ and $\mathbf{V}$ into their Householder reflectors and use them to initialize our projector. Such an initialization scheme leverages the statistics of the original weight matrix, which might give our projector a good starting point and improve the performance (see the ablation study of Sec. D.3 in the supplementary). 


\noindent\textbf{Acceleration.} The accumulated product of elementary Householder matrices can be accelerated via the theorem: 

\begin{thm}[Compact WY representation~\cite{bischof1987wy}]
For any accumulation of $m$ Householder matrices $\mathbf{H}_{1}{\dots}\mathbf{H}_{m}$, there exists $\mathbf{W}{,}\mathbf{Y}{\in}\mathrm{R}^{{d}{\times}m}$ such that $\mathbf{I}{-}2\mathbf{W}\mathbf{Y}^{T}{=}\mathbf{H}_{1}{\dots}\mathbf{H}_{m}$ where computing $\mathbf{W}$ and $\mathbf{Y}$ takes $O(dm^2)$ time and $m$ sequential Householder multiplications. 
\end{thm}

This theorem indicates the possibility to repeatedly split the accumulation $\mathbf{H}_{1}\mathbf{H}_{2}\dots\mathbf{H}_{m}$ into multiple sub-sequences until irreducible. Then these sub-sequences can be computed in parallel and gradually merged. As revealed in the ablation of Sec. D.3 in the supplementary, this technique could fully exploit the parallel computational power of GPUs and greatly speed up the aggregation, particularly in our case where the projector dimension is large.

\begin{figure}[tbp]
    \centering
    \includegraphics[width=0.99\linewidth]{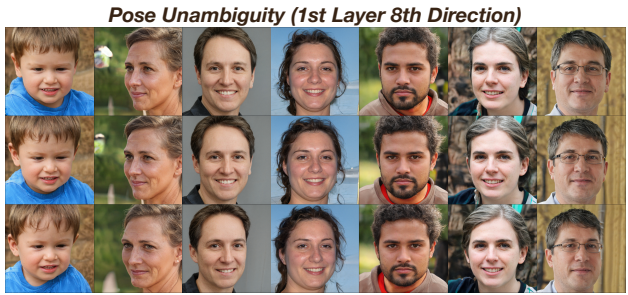}
    \caption{Interpretable directions identified by our method are semantically consistent among different samples.}
    \label{fig:unambiguity}
\end{figure}

\begin{table*}[t]
    \centering
    \begin{minipage}{0.24\linewidth}
    \resizebox{0.99\linewidth}{!}{
    \begin{tabular}{c|c|c|c|c|c|c}
    \toprule
         & Identity & Pose & Age & Gender & Glasses & Smile \\
    \midrule
        Identity &\cellcolor{purple!20}\textbf{0.65} &0.24 &0.03 &0.04 & 0.01 &0.03 \\
        Pose &0.11&\cellcolor{purple!20}\textbf{0.57} &0.04 &0.04 &0.11 &0.13  \\
        Age &0.02&0.05 &\cellcolor{purple!20}\textbf{0.67} &0.03 &0.19 &0.03  \\
        Gender &0.03&0.32 &0.02 &\cellcolor{purple!20}\textbf{0.52} &0.10 &0.00 \\
        Glasses &0.01&0.08 &0.00 &0.02 &\cellcolor{purple!20}\textbf{0.88} &0.01  \\
        Smile &0.02&0.01 &0.01 &0.00 &0.01 &\cellcolor{purple!20}\textbf{0.95} \\
    \bottomrule
    \end{tabular}
    }
    \vspace{-3mm}
    \captionof{table}*{(a) Our method}
    \end{minipage}
    \begin{minipage}{0.24\linewidth}
    \resizebox{0.99\linewidth}{!}{
    \begin{tabular}{c|c|c|c|c|c|c}
    \toprule
         & Identity & Pose & Age & Gender & Glasses & Smile \\
    \midrule
        Identity &\cellcolor{purple!20}\textbf{0.56} &0.06 &0.05 &0.02 & 0.05 &0.26 \\
        Pose &0.44&\cellcolor{purple!20}\textbf{0.48} &0.05 &0.01 &0.01 &0.00  \\
        Age &\cellcolor{purple!20}\textbf{0.43}&0.27 &0.22 &0.05 &0.03 &0.01  \\
        Gender &0.33&0.18 &0.04 &\cellcolor{purple!20}\textbf{0.40} &0.04 &0.02 \\
        Glasses &\cellcolor{purple!20}\textbf{0.27}&0.14 &0.04 &0.10 &0.23 &0.22  \\
        Smile &0.20&0.08 &0.09 &0.00 &0.03 &\cellcolor{purple!20}\textbf{0.61} \\
    \bottomrule
    \end{tabular}
    }
    \vspace{-3mm}
    \captionof{table}*{(b) SeFa}
    \end{minipage}
    \begin{minipage}{0.24\linewidth}
    \resizebox{0.99\linewidth}{!}{
    \begin{tabular}{c|c|c|c|c|c|c}
    \toprule
         & Identity & Pose & Age & Gender & Glasses & Smile \\
    \midrule
       Identity &\cellcolor{purple!20}\textbf{0.51} &0.27 &0.04 &0.05 & 0.03 &0.10 \\
        Pose &\cellcolor{purple!20}\textbf{0.39}&0.35 &0.05 &0.02 &0.07 &0.11  \\
        Age &0.26&0.14 &\cellcolor{purple!20}\textbf{0.32} &0.10 &0.05 &0.12  \\
        Gender &0.20&0.06 &\cellcolor{purple!20}\textbf{0.34} &0.29 &0.04 &0.07 \\
        Glasses &0.18&0.06 &0.12 &0.07 &\cellcolor{purple!20}\textbf{0.42} &0.14  \\
        Smile &0.13&0.05 &0.07 &0.02 &0.03 &\cellcolor{purple!20}\textbf{0.70} \\
    \bottomrule
    \end{tabular}
    }
    \vspace{-3mm}
    \captionof{table}*{(c) OrJaR}
    \end{minipage}
    \begin{minipage}{0.24\linewidth}
    \resizebox{0.99\linewidth}{!}{
    \begin{tabular}{c|c|c|c|c|c|c}
    \toprule
         & Identity & Pose & Age & Gender & Glasses & Smile \\
    \midrule
        Identity &\cellcolor{purple!20}\textbf{0.54} &0.25 &0.04 &0.06 & 0.02 &0.08 \\
        Pose &\cellcolor{purple!20}\textbf{0.42}&0.30 &0.08 &0.04 &0.05 &0.12  \\
        Age &0.21&0.15 &\cellcolor{purple!20}\textbf{0.28} &0.11 &0.08 &0.17  \\
        Gender &\cellcolor{purple!20}\textbf{0.27}&0.07 &0.23 &0.25 &0.04 &0.13 \\
        Glasses &0.15&0.12 &0.16 &0.04 &\cellcolor{purple!20}\textbf{0.37} &0.16  \\
        Smile &0.11&0.09 &0.12 &0.01 &0.01 &\cellcolor{purple!20}\textbf{0.66} \\
    \bottomrule
    \end{tabular}
    }
     \vspace{-3mm}
    \captionof{table}*{(d) HP}
    \end{minipage}
    \vspace{-3mm}
    \caption{The $l_{1}$ normalized attribute correlations based on $2K$ same samples generated by GAN inversion. }
    \label{tab:att_corr}
\end{table*}

\begin{figure}[h]
    \centering
    \includegraphics[width=0.99\linewidth]{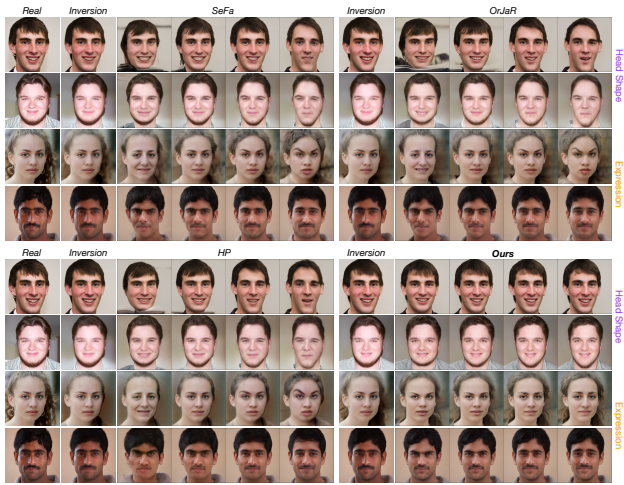}
    \caption{Qualitative comparisons of the same samples.}
    \label{fig:gan_inversion}
\end{figure}

\begin{figure*}[htbp]
    \centering
    \includegraphics[width=0.85\linewidth]{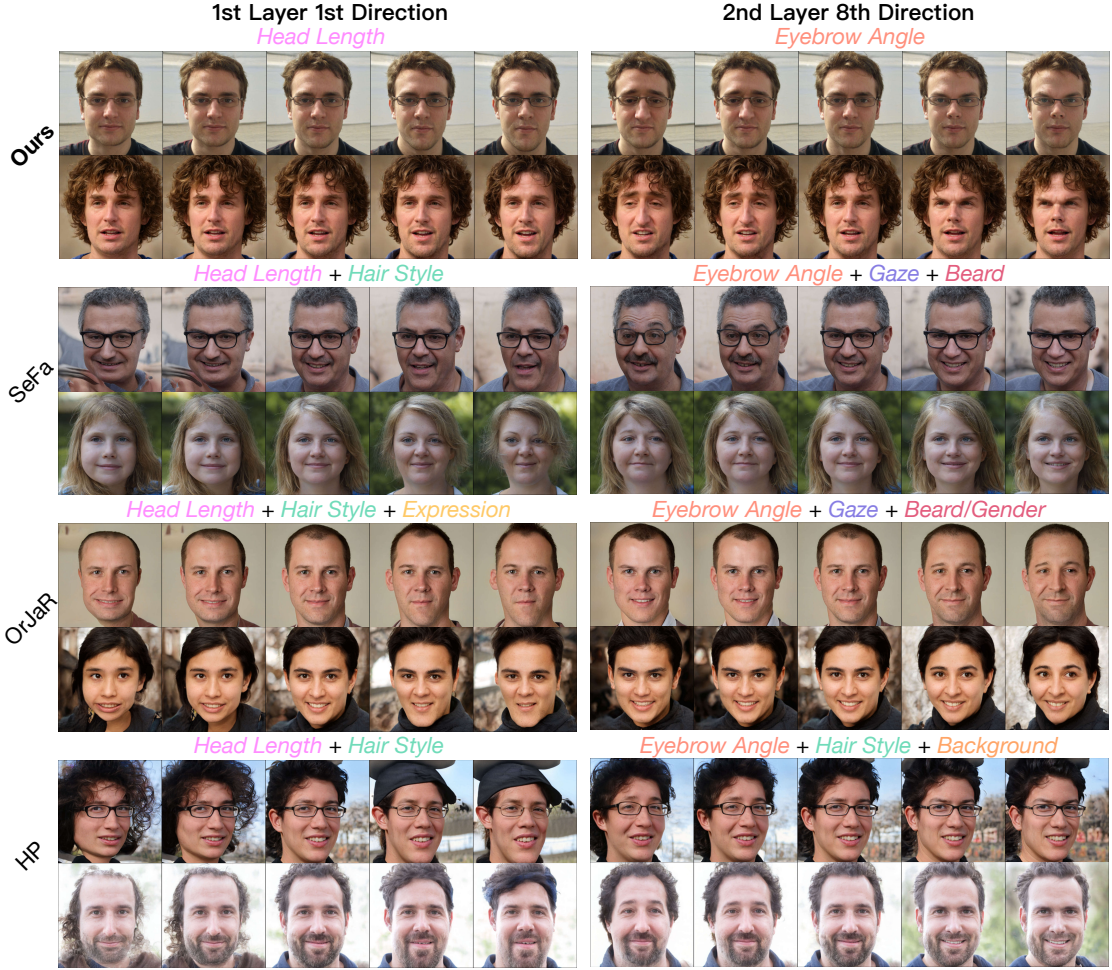}
    \caption{Exemplary qualitative comparison of two different semantics on FFHQ~\cite{kazemi2014one} with StyleGAN2~\cite{karras2020analyzing}. Our Householder Projector can precisely control the image attributes without changing the face identity. The direction index denotes the index of eigenvectors.}
    \label{fig:com_ffhq}
    \vspace{-0.4cm}
\end{figure*}


\noindent\textbf{Implementation in StyleGANs.} Fig.~\ref{fig:model} depicts how our Householder Projector modifies the original StyleGAN architectures. The projector used in the weight modulation module is represented by our proposed projector. The weight matrix is thus endowed with low-rank orthogonal properties. We insert the proposed projector at every layer of StyleGAN2 and every four layers of StyleGAN3. Since StyleGAN3 has $15$ intermediate layers, the adjacent layers have very similar and even repeated semantics. Therefore, we choose to integrate our projector every few layers to obtain interpretable directions of different semantics.

\section{Experiments}
In this section, we first introduce the experimental setup, followed by the quantitative and qualitative evaluation. \textbf{\textit{We defer the full ablation studies to Sec. D of supplementary.}}

\subsection{Setup}

\noindent\textbf{Models.} We evaluate our Householder Projector on StyleGAN2~\cite{karras2020analyzing} and StyleGAN3~\cite{karras2021alias}, \emph{i.e.,} the challenging \emph{state-of-the-art} GAN backbones in the field of computer vision. 

\noindent\textbf{Datasets and Baselines.} For StyleGAN2, we conduct experiments on FFHQ~\cite{kazemi2014one}, LSUN Church~\cite{yu2015lsun}, and LSUN Cat~\cite{yu2015lsun}. The experiments of StyleGAN3 are performed on SHHQ~\cite{fu2022styleganhuman}, MetFaces~\cite{karras2020training} and AFHQv2~\cite{choi2020stargan}. We mainly compare our method with representative unsupervised latent semantics discovery approaches, including SeFa~\cite{shen2021closed}, Orthogonal Jacobian Regularization (OrJaR)~\cite{wei2021orthogonal}, and Hessian Penalty (HP)~\cite{peebles2020hessian}. SeFa~\cite{shen2021closed} can be directly applied to pre-trained models, while OrJaR and HP need extra fine-tuning or training from scratch due to the regularization. 



\noindent\textbf{Metrics.} We conduct quantitative evaluation using (1) \textbf{Fréchet Inception Distance (FID)}~\cite{heusel2017gans}, (2) \textbf{Perceptual Path Length (PPL)}~\cite{karras2019style}, (3) \textbf{Perceptual Interpretable Path Length (PIPL)}, and (4) \textbf{Face Attribute Correlation}. FID aims to measure the image quality and diversity by computing the distance between the real and fake distributions, and PPL is designed to assess the perceptual smoothness of the latent space where the smoothness can reflect the disentanglement ability. Our proposed PIPL is a natural modification of PPL by adapting the latent manipulation from random interpolation-based perturbations to vector-based perturbations using interpretable directions. Compared with PPL, our PIPL can better measure the latent space smoothness when the latent code is perturbed along with specific vectors, which suits more vector-based semantic discovery methods like SeFa~\cite{shen2021closed} and ours. Furthermore, for StyleGAN2 trained on FFHQ, we rely on pre-trained face attribute estimators to compute the correlation coefficient between the traversal steps and the face attributes.  Besides the above four metrics, we also assess the disentanglement through visual observation. We defer the details of used datasets and metrics to Sec. C of the supplementary.




\noindent\textbf{Implementation Details.} We adopt the widely used Pytorch implementation of StyleGAN2\footnote{https://github.com/rosinality/stylegan2-pytorch} and convert the official TensorFlow pre-trained models into PyTorch for FFHQ~\cite{kazemi2014one}, LSUN Church~\cite{yu2015lsun}, and LSUN Cat~\cite{yu2015lsun}. For StyleGAN3, we use the official code and pre-trained models of AFHQv2~\cite{choi2020stargan} and MetFaces~\cite{karras2020training}. As for SHHQ~\cite{fu2022styleganhuman}, we also use the official pre-trained model\footnote{https://github.com/stylegan-human/StyleGAN-Human}. Following the original optimization strategy, we finetune all the parameters of the pre-trained generator and discriminator within $1\%$ of the total training steps (kimgs for StyleGAN3). For instance, the fine-tuning process takes $5K$ steps for StyleGAN2 with FFHQ and $250$ kimgs for StyleGAN3 with AFHQ. The fine-tuning time is thus very limited due to the small number of training steps. To give concrete examples, fine-tuning models on FFHQ and AFHQ takes $1.5$ and $2.5$ hours, respectively. The rank $N$ of $\mathbf{S}$ is set to 10 for all experiments based on our empirical observation of the number of semantics of StyleGANs. Our editing strength is set the same as SeFa. We use 4 RTX A6000 GPUs for the training. For the comparison fairness, the baseline methods OrJaR~\cite{wei2021orthogonal} and HP~\cite{peebles2020hessian} are also fine-tuned with the same steps.


\begin{table}[h]
\vspace{-0.2cm}
    \centering
    \resizebox{0.9\linewidth}{!}{
    \begin{tabular}{c|c|ccc}
    \toprule
        \textbf{Datasets}& \textbf{Methods} &\textbf{FID ($\downarrow$)}  & \textbf{PPL ($\downarrow$)} & \textbf{PIPL ($\downarrow$)}\\
        \midrule
        \multirow{4}*{\makecell[c]{\textbf{FFHQ}~\cite{kazemi2014one}\\ $1024\times1024$}} 
         &SeFa~\cite{shen2021closed} &4.48 &1579.76 &0.227 \\ 
         &OrJaR~\cite{wei2021orthogonal} &4.51 &987.80 & 0.204\\ 
         &HP~\cite{peebles2020hessian} &4.66 &993.17 & 0.207 \\ 
         \rowcolor{purple!20}\cellcolor{white}&Ours &\textbf{4.40} & \textbf{966.23} &\textbf{0.141}\\ 
         \midrule
        \multirow{4}*{\makecell[c]{\textbf{LSUN Church}~\cite{yu2015lsun}\\ $256\times256$}} &SeFa~\cite{shen2021closed} &4.61 &530.68 & 0.069\\
         &OrJaR~\cite{wei2021orthogonal} & 3.77 & 474.77 & 0.065\\
         &HP~\cite{peebles2020hessian} & 3.78 & 486.93 & 0.058\\
          \rowcolor{purple!20}\cellcolor{white}&Ours & \textbf{3.72}& \textbf{457.52} & \textbf{0.030}\\ 
         \midrule
        \multirow{4}*{\makecell[c]{\textbf{LSUN Cat}~\cite{yu2015lsun}\\ $256\times256$}} &SeFa~\cite{shen2021closed} &8.37 & 722.24 & 0.141 \\
         &OrJaR~\cite{wei2021orthogonal} & 8.39 & 562.98 & 0.134\\
         &HP~\cite{peebles2020hessian} & \textbf{8.31} & 573.71 & 0.136\\
         \rowcolor{purple!20}\cellcolor{white}&Ours & 8.46 & \textbf{526.26}&\textbf{0.057}\\ 
    \bottomrule
    \end{tabular}
    }
    \caption{Evaluation results on StyleGAN2.}
    \label{tab:results_stylegan2}
    \vspace{-0.4cm}
\end{table}

\begin{figure*}[t]
    \centering
    \includegraphics[width=0.85\linewidth]{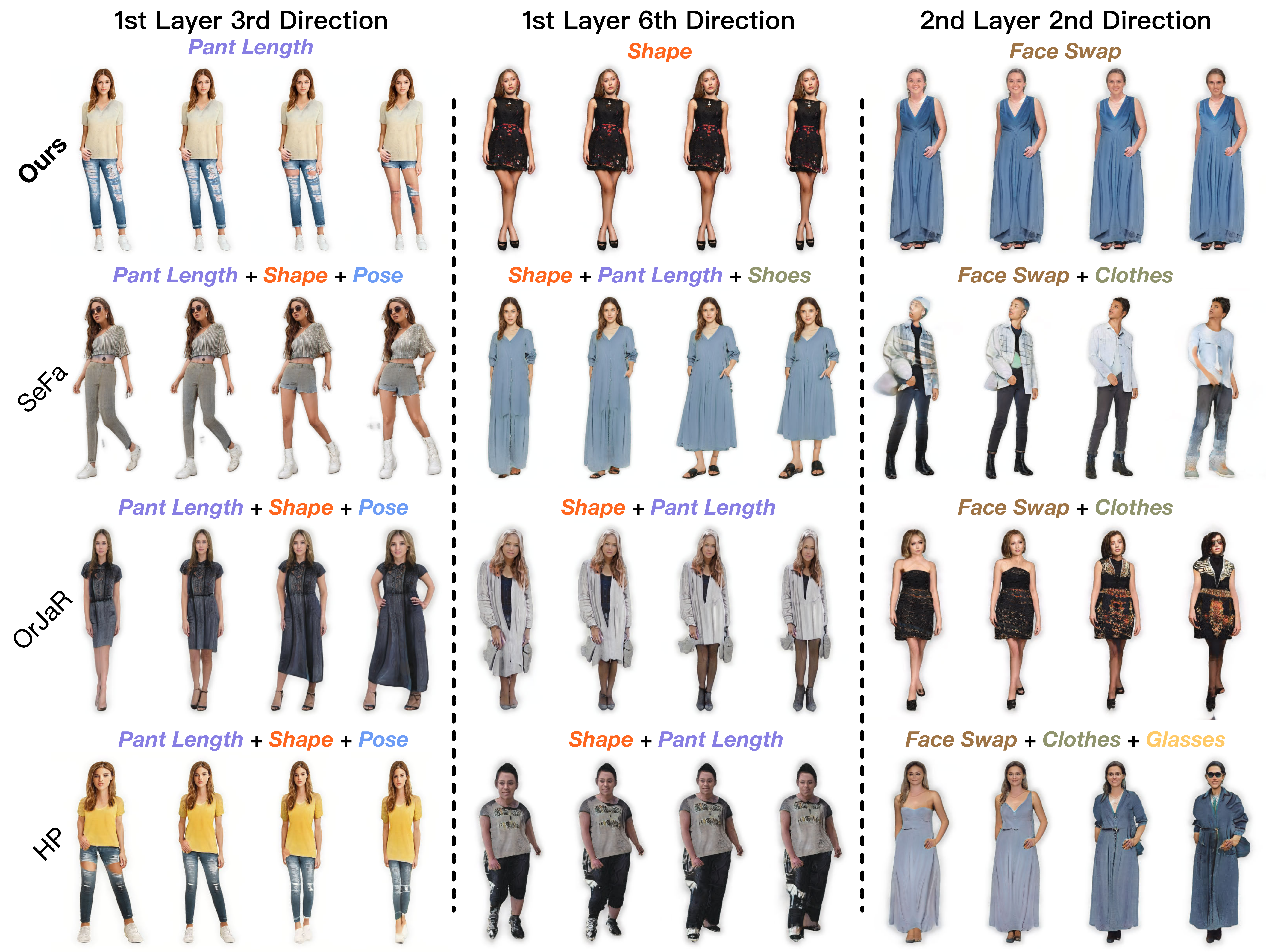}
    \caption{Exemplary visual comparison of three different semantics on SHHQ~\cite{fu2022styleganhuman} with StyleGAN3~\cite{karras2021alias}. Our method is able to mine more disentangled interpretable directions and have more precise control on the attributes. The direction index denotes the index of eigenvectors.}
    \label{fig:com_human}
    \vspace{-0.4cm}
\end{figure*}

\subsection{Qualitative Evaluation and Discussion}
\label{sec:quali}

\noindent\textbf{Diverse and Precise Attributes.} Fig.~\ref{fig:diversity} exhibits some semantic attributes discovered by our Householder Projector on all the datasets. Our method mines a diverse set of disentangled semantics, enabling a wide range of attribute manipulation. Take as an example the first row of attributes discovered on FFHQ~\cite{kazemi2014one}. The left columns present diverse high-level semantic concepts such as ``Pose'' and ``Shape'', while the right columns show low-level attributes such as ``Color'' and ``Lighting''. This hierarchy also meets the same trend of original StyleGANs. The manipulation of the diverse and highly-disentangled semantics would give users more precise control on the image generation process. 


\noindent\textbf{Semantic Unambiguity.} Importantly, our interpretable directions are unambiguous to different samples. As shown in Fig.~\ref{fig:unambiguity}, the image variations manipulated by our discovered directions all correspond to the same semantic attribute, \emph{i.e.}, the head pose. The other non-target attributes are untouched, such as the background and face identity. 


\noindent\textbf{Comparison against Other Methods.} Fig.~\ref{fig:com_ffhq} and Fig.~\ref{fig:com_human} compare the latent traversal of some directions against other baselines on FFHQ~\cite{kazemi2014one} and SHHQ~\cite{fu2022styleganhuman}, respectively. All the methods can discover similar attributes in the same layer, such as the head length in Fig.~\ref{fig:com_ffhq} left. However, the baselines suffer from entangled semantics and the other attributes vary during the traversal, such as hairstyle for SeFA~\cite{shen2021closed} and OrJaR~\cite{wei2021orthogonal}, and expression for HP~\cite{peebles2020hessian}. By contrast, our method is able to discover more precise semantics and preserve other non-target attributes.


\noindent\textbf{Comparison of Same Samples.} To better compare the qualitative disentanglement performance, we further use a GAN inversion technique (PTI~\cite{roich2022pivotal}) to create nearly the same images from FFHQ for different methods. Fig.~\ref{fig:gan_inversion} displays some qualitative comparisons. For the same samples, our method still has more precise attribute control.

\noindent\textbf{Local Editing Applications.} With precise attribute control, our method is even able to edit local regions by simply performing latent traversal. Fig.~\ref{fig:local} displays such a use case. Our method achieves very competitive performance against local editing methods such as the recent ReSeFa~\cite{zhu2022region}. In addition, our method is free of extra bounding box as input.


Please refer to Sec. E of the supplementary for more visualizations about comparisons on other datasets, and semantic diversity/unambiguity/hierarchy.


\subsection{Quantitative Evaluation}


\noindent\textbf{StyleGAN2 Results.} Table~\ref{tab:results_stylegan2} presents the quantitative evaluation results on FFHQ~\cite{kazemi2014one}, LSUN Church~\cite{yu2015lsun} and LSUN Cat~\cite{yu2015lsun} datasets. Our proposed Householder Projector outperforms other baselines in terms of both PPL and PIPL. This demonstrates that our method has a smoother and more structured latent space, which corresponds to more disentangled representations.
In particular, our approach surpasses SeFa~\cite{shen2021closed} by a large margin, which indicates the benefit of enforcing low-rank orthogonality to the projection matrix. Compared with the \textit{soft} orthogonality regularization used in OrJaR~\cite{wei2021orthogonal} and HP~\cite{peebles2020hessian}, the \textit{hard} orthogonality of our projector also has an advantage in latent smoothness due to the strict orthogonality preservation and the additional low-rank constraint. Moreover, our FID score is also very competitive, implying that our method could simultaneously improve the disentanglement performance while keeping the quality of generated images unharmed. For the attribute correlation, we also use GAN inversion to create a dataset of $2K$ identical images for each method. Table~\ref{tab:att_corr} compares the correlation results on FFHQ. Our method outperforms other unsupervised baselines and preserves the attribute well in particular. 



\begin{table}[h]
    \centering
    \resizebox{0.9\linewidth}{!}{
    \begin{tabular}{c|c|ccc}
    \toprule
        \textbf{Datasets}& \textbf{Methods} &\textbf{FID ($\downarrow$)}  & \textbf{PPL ($\downarrow$)} & \textbf{PIPL ($\downarrow$)}\\
        \midrule
        \multirow{4}*{\makecell[c]{\textbf{MetFaces}~\cite{karras2020training}\\ $1024\times1024$}} &SeFa~\cite{shen2021closed} &\textbf{15.33} & 5626.31 & 2.991 \\
         &OrJaR~\cite{wei2021orthogonal} & 17.55 & 5754.44 & 3.700 \\
         &HP~\cite{peebles2020hessian} & 17.32 & 5323.27 & 3.465  \\
         \rowcolor{purple!20}\cellcolor{white}&Ours &16.89 & \textbf{4192.52} & \textbf{0.099} \\
         \midrule
        \multirow{4}*{\makecell[c]{\textbf{AFHQv2}~\cite{choi2020stargan}\\ $512\times512$}} &SeFa~\cite{shen2021closed} &\textbf{4.40} &2193.74& 0.470 \\
         &OrJaR~\cite{wei2021orthogonal} &5.45 &2103.47 & 0.440 \\
         &HP~\cite{peebles2020hessian} &5.33 &2198.46 & 0.463 \\
          \rowcolor{purple!20}\cellcolor{white}&Ours &4.98 & \textbf{2052.38} & \textbf{0.070} \\
         \midrule
        \multirow{4}*{\makecell[c]{\textbf{SHHQ}~\cite{fu2022styleganhuman}\\ $512\times256$}} &SeFa~\cite{shen2021closed} &\textbf{2.54} & 1621.07 &0.370 \\
         &OrJaR~\cite{wei2021orthogonal} &4.78 &1614.56 & 0.245 \\ 
         &HP~\cite{peebles2020hessian} &5.38 &1648.74 & 0.216 \\ 
         \rowcolor{purple!20}\cellcolor{white}&Ours & 4.17 &\textbf{1549.01} & \textbf{0.119} \\
    \bottomrule
    \end{tabular}
    }
    \caption{Evaluation results on StyleGAN3.}
    \label{tab:results_stylegan3}
    \vspace{-0.2cm}
\end{table}

\noindent\textbf{StyleGAN3 Results.} Table~\ref{tab:results_stylegan3} compares the performance of our method against other baselines on MetFaces~\cite{karras2020training}, AFHQv2~\cite{choi2020stargan}, and SHHQ~\cite{fu2022styleganhuman} datasets. The results are very coherent with those on StyleGAN2: our Householder Projector improves the latent space smoothness without harming the image fidelity. Our method as well as the two baselines that involve fine-tuning have slightly worse FID than the original StyleGAN3. This might stem from the fact that due to the limited computational resources, our used batch size ($16$) is actually smaller than the original setting of StyleGAN3 ($32$). As revealed in the ablation study of Sec. D.2 of the supplementary, the hyper-parameter batch size has a substantial impact on FID. We expect that increasing the batch size would further boost the image quality of our method and lead to a more competitive FID score. 








\section{Conclusion}
This paper proposes a general and flexible low-rank orthogonal matrix representation coined as Householder Projector for unsupervised latent semantics discovery of generative models. The proposed method endows the projection matrix with low-rank orthogonality. This superiority helps pre-trained GANs to achieve precise and diverse semantics control within limited fine-tuning steps. Extensive experiments of StyleGANs on various benchmarks demonstrate that our method could simultaneously identify the disentangled attributes while maintaining image fidelity.

{\small
\bibliographystyle{ieee_fullname}
\bibliography{egbib}

\begin{thebibliography}{10}\itemsep=-1pt

\bibitem{abdal2021styleflow}
Rameen Abdal, Peihao Zhu, Niloy~J Mitra, and Peter Wonka.
\newblock Styleflow: Attribute-conditioned exploration of stylegan-generated
  images using conditional continuous normalizing flows.
\newblock {\em ACM TOG}, 2021.

\bibitem{bau2020semantic}
David Bau, Hendrik Strobelt, William Peebles, Jonas Wulff, Bolei Zhou, Jun-Yan
  Zhu, and Antonio Torralba.
\newblock Semantic photo manipulation with a generative image prior.
\newblock {\em ACM TOG}, 2020.

\bibitem{bau2019gan}
David Bau, Jun-Yan Zhu, Hendrik Strobelt, Bolei Zhou, Joshua~B Tenenbaum,
  William~T Freeman, and Antonio Torralba.
\newblock Gan dissection: Visualizing and understanding generative adversarial
  networks.
\newblock In {\em ICLR}, 2019.

\bibitem{bischof1987wy}
Christian Bischof and Charles Van~Loan.
\newblock The wy representation for products of householder matrices.
\newblock {\em SIAM Journal on Scientific and Statistical Computing},
  8(1):s2--s13, 1987.

\bibitem{brock2019large}
Andrew Brock, Jeff Donahue, and Karen Simonyan.
\newblock Large scale gan training for high fidelity natural image synthesis.
\newblock {\em ICLR}, 2019.

\bibitem{Chan2022}
Eric~R. Chan, Connor~Z. Lin, Matthew~A. Chan, Koki Nagano, Boxiao Pan,
  Shalini~De Mello, Orazio Gallo, Leonidas Guibas, Jonathan Tremblay, Sameh
  Khamis, Tero Karras, and Gordon Wetzstein.
\newblock Efficient geometry-aware {3D} generative adversarial networks.
\newblock In {\em CVPR}, 2022.

\bibitem{chen2016infogan}
Xi Chen, Yan Duan, Rein Houthooft, John Schulman, Ilya Sutskever, and Pieter
  Abbeel.
\newblock Infogan: Interpretable representation learning by information
  maximizing generative adversarial nets.
\newblock {\em NeurIPS}, 2016.

\bibitem{chen2022exploring}
Zikun Chen, Ruowei Jiang, Brendan Duke, Han Zhao, and Parham Aarabi.
\newblock Exploring gradient-based multi-directional controls in gans.
\newblock {\em ECCV}, 2022.

\bibitem{choi2022not}
Jaewoong Choi, Junho Lee, Changyeon Yoon, Jung~Ho Park, Geonho Hwang, and
  Myungjoo Kang.
\newblock Do not escape from the manifold: Discovering the local coordinates on
  the latent space of gans.
\newblock {\em ICLR}, 2022.

\bibitem{choi2020stargan}
Yunjey Choi, Youngjung Uh, Jaejun Yoo, and Jung-Woo Ha.
\newblock Stargan v2: Diverse image synthesis for multiple domains.
\newblock In {\em CVPR}, 2020.

\bibitem{collins2020editing}
Edo Collins, Raja Bala, Bob Price, and Sabine Susstrunk.
\newblock Editing in style: Uncovering the local semantics of gans.
\newblock In {\em CVPR}, 2020.

\bibitem{deng2020disentangled}
Yu Deng, Jiaolong Yang, Dong Chen, Fang Wen, and Xin Tong.
\newblock Disentangled and controllable face image generation via 3d
  imitative-contrastive learning.
\newblock In {\em CVPR}, 2020.

\bibitem{doosti2020hope}
Bardia Doosti, Shujon Naha, Majid Mirbagheri, and David~J Crandall.
\newblock Hope-net: A graph-based model for hand-object pose estimation.
\newblock In {\em CVPR}, 2020.

\bibitem{fu2022styleganhuman}
Jianglin Fu, Shikai Li, Yuming Jiang, Kwan-Yee Lin, Chen Qian, Chen-Change Loy,
  Wayne Wu, and Ziwei Liu.
\newblock Stylegan-human: A data-centric odyssey of human generation.
\newblock {\em ECCV}, 2022.

\bibitem{goetschalckx2019ganalyze}
Lore Goetschalckx, Alex Andonian, Aude Oliva, and Phillip Isola.
\newblock Ganalyze: Toward visual definitions of cognitive image properties.
\newblock In {\em ICCV}, 2019.

\bibitem{goodfellow2014generative}
Ian Goodfellow, Jean Pouget-Abadie, Mehdi Mirza, Bing Xu, David Warde-Farley,
  Sherjil Ozair, Aaron Courville, and Yoshua Bengio.
\newblock Generative adversarial nets.
\newblock {\em NeurIPS}, 2014.

\bibitem{gu2021stylenerf}
Jiatao Gu, Lingjie Liu, Peng Wang, and Christian Theobalt.
\newblock Stylenerf: A style-based 3d-aware generator for high-resolution image
  synthesis.
\newblock {\em ICLR}, 2022.

\bibitem{harkonen2020ganspace}
Erik H{\"a}rk{\"o}nen, Aaron Hertzmann, Jaakko Lehtinen, and Sylvain Paris.
\newblock Ganspace: Discovering interpretable gan controls.
\newblock {\em NeurIPS}, 2020.

\bibitem{he2021eigengan}
Zhenliang He, Meina Kan, and Shiguang Shan.
\newblock Eigengan: Layer-wise eigen-learning for gans.
\newblock In {\em ICCV}, 2021.

\bibitem{heusel2017gans}
Martin Heusel, Hubert Ramsauer, Thomas Unterthiner, Bernhard Nessler, and Sepp
  Hochreiter.
\newblock Gans trained by a two time-scale update rule converge to a local nash
  equilibrium.
\newblock {\em NeurIPS}, 2017.

\bibitem{householder1958unitary}
Alston~S Householder.
\newblock Unitary triangularization of a nonsymmetric matrix.
\newblock {\em Journal of the ACM (JACM)}, 5(4):339--342, 1958.

\bibitem{jahanian2020steerability}
Ali Jahanian, Lucy Chai, and Phillip Isola.
\newblock On the" steerability" of generative adversarial networks.
\newblock In {\em ICLR}, 2020.

\bibitem{kappiyath2022self}
Adarsh Kappiyath, Silpa~Vadakkeeveetil Sreelatha, and S Sumitra.
\newblock Self-supervised enhancement of latent discovery in gans.
\newblock In {\em AAAI}, 2022.

\bibitem{karkkainen2021fairface}
Kimmo Karkkainen and Jungseock Joo.
\newblock Fairface: Face attribute dataset for balanced race, gender, and age
  for bias measurement and mitigation.
\newblock In {\em WACV}, 2021.

\bibitem{karmali2022hierarchical}
Tejan Karmali, Rishubh Parihar, Susmit Agrawal, Harsh Rangwani, Varun Jampani,
  Maneesh Singh, and R~Venkatesh Babu.
\newblock Hierarchical semantic regularization of latent spaces in stylegans.
\newblock In {\em ECCV}, 2022.

\bibitem{karras2018progressive}
Tero Karras, Timo Aila, Samuli Laine, and Jaakko Lehtinen.
\newblock Progressive growing of gans for improved quality, stability, and
  variation.
\newblock {\em ICLR}, 2018.

\bibitem{karras2020training}
Tero Karras, Miika Aittala, Janne Hellsten, Samuli Laine, Jaakko Lehtinen, and
  Timo Aila.
\newblock Training generative adversarial networks with limited data.
\newblock {\em NeurIPS}, 2020.

\bibitem{karras2021alias}
Tero Karras, Miika Aittala, Samuli Laine, Erik H{\"a}rk{\"o}nen, Janne
  Hellsten, Jaakko Lehtinen, and Timo Aila.
\newblock Alias-free generative adversarial networks.
\newblock {\em NeurIPS}, 2021.

\bibitem{karras2019style}
Tero Karras, Samuli Laine, and Timo Aila.
\newblock A style-based generator architecture for generative adversarial
  networks.
\newblock In {\em CVPR}, 2019.

\bibitem{karras2020analyzing}
Tero Karras, Samuli Laine, Miika Aittala, Janne Hellsten, Jaakko Lehtinen, and
  Timo Aila.
\newblock Analyzing and improving the image quality of stylegan.
\newblock In {\em CVPR}, 2020.

\bibitem{kazemi2014one}
Vahid Kazemi and Josephine Sullivan.
\newblock One millisecond face alignment with an ensemble of regression trees.
\newblock In {\em CVPR}, 2014.

\bibitem{kim2021exploiting}
Hyunsu Kim, Yunjey Choi, Junho Kim, Sungjoo Yoo, and Youngjung Uh.
\newblock Exploiting spatial dimensions of latent in gan for real-time image
  editing.
\newblock In {\em CVPR}, 2021.

\bibitem{kwon2021diagonal}
Gihyun Kwon and Jong~Chul Ye.
\newblock Diagonal attention and style-based gan for content-style
  disentanglement in image generation and translation.
\newblock In {\em CVPR}, 2021.

\bibitem{lehoucq1996computation}
Richard~B Lehoucq.
\newblock The computation of elementary unitary matrices.
\newblock {\em ACM Transactions on Mathematical Software (TOMS)},
  22(4):393--400, 1996.

\bibitem{li2021surrogate}
Minjun Li, Yanghua Jin, and Huachun Zhu.
\newblock Surrogate gradient field for latent space manipulation.
\newblock In {\em CVPR}, 2021.

\bibitem{ling2021editgan}
Huan Ling, Karsten Kreis, Daiqing Li, Seung~Wook Kim, Antonio Torralba, and
  Sanja Fidler.
\newblock Editgan: High-precision semantic image editing.
\newblock {\em NeurIPS}, 2021.

\bibitem{mathiasen2020if}
Alexander Mathiasen, Frederik Hvilsh{\o}j, Jakob R{\o}dsgaard~J{\o}rgensen,
  Anshul Nasery, and Davide Mottin.
\newblock What if neural networks had svds?
\newblock {\em NeurIPS}, 2020.

\bibitem{mhammedi2017efficient}
Zakaria Mhammedi, Andrew Hellicar, Ashfaqur Rahman, and James Bailey.
\newblock Efficient orthogonal parametrisation of recurrent neural networks
  using householder reflections.
\newblock In {\em ICML}, 2017.

\bibitem{oldfield2021tensor}
James Oldfield, Markos Georgopoulos, Yannis Panagakis, Mihalis~A Nicolaou, and
  Ioannis Patras.
\newblock Tensor component analysis for interpreting the latent space of gans.
\newblock {\em BMVC}, 2021.

\bibitem{oldfield2022panda}
James Oldfield, Christos Tzelepis, Yannis Panagakis, Mihalis~A Nicolaou, and
  Ioannis Patras.
\newblock Panda: Unsupervised learning of parts and appearances in the feature
  maps of gans.
\newblock {\em ICLR}, 2023.

\bibitem{or2022stylesdf}
Roy Or-El, Xuan Luo, Mengyi Shan, Eli Shechtman, Jeong~Joon Park, and Ira
  Kemelmacher-Shlizerman.
\newblock Stylesdf: High-resolution 3d-consistent image and geometry
  generation.
\newblock In {\em CVPR}, 2022.

\bibitem{patashnik2021styleclip}
Or Patashnik, Zongze Wu, Eli Shechtman, Daniel Cohen-Or, and Dani Lischinski.
\newblock Styleclip: Text-driven manipulation of stylegan imagery.
\newblock In {\em ICCV}, 2021.

\bibitem{peebles2020hessian}
William Peebles, John Peebles, Jun-Yan Zhu, Alexei Efros, and Antonio Torralba.
\newblock The hessian penalty: A weak prior for unsupervised disentanglement.
\newblock In {\em ECCV}. Springer, 2020.

\bibitem{plumerault2020controlling}
Antoine Plumerault, Herv{\'e}~Le Borgne, and C{\'e}line Hudelot.
\newblock Controlling generative models with continuous factors of variations.
\newblock {\em ICLR}, 2020.

\bibitem{radford2015unsupervised}
Alec Radford, Luke Metz, and Soumith Chintala.
\newblock Unsupervised representation learning with deep convolutional
  generative adversarial networks.
\newblock {\em ICLR}, 2016.

\bibitem{ren2022learning}
Xuanchi Ren, Tao Yang, Yuwang Wang, and Wenjun Zeng.
\newblock Learning disentangled representation by exploiting pretrained
  generative models: A contrastive learning view.
\newblock In {\em ICLR}, 2022.

\bibitem{roich2022pivotal}
Daniel Roich, Ron Mokady, Amit~H Bermano, and Daniel Cohen-Or.
\newblock Pivotal tuning for latent-based editing of real images.
\newblock {\em ACM TOG}, 2022.

\bibitem{sauer2022stylegan}
Axel Sauer, Katja Schwarz, and Andreas Geiger.
\newblock Stylegan-xl: Scaling stylegan to large diverse datasets.
\newblock In {\em ACM SIGGRAPH}, 2022.

\bibitem{shen2020interpreting}
Yujun Shen, Jinjin Gu, Xiaoou Tang, and Bolei Zhou.
\newblock Interpreting the latent space of gans for semantic face editing.
\newblock In {\em CVPR}, 2020.

\bibitem{shen2021closed}
Yujun Shen and Bolei Zhou.
\newblock Closed-form factorization of latent semantics in gans.
\newblock In {\em CVPR}, 2021.

\bibitem{shi2022semanticstylegan}
Yichun Shi, Xiao Yang, Yangyue Wan, and Xiaohui Shen.
\newblock Semanticstylegan: Learning compositional generative priors for
  controllable image synthesis and editing.
\newblock In {\em CVPR}, 2022.

\bibitem{skorokhodov2022stylegan}
Ivan Skorokhodov, Sergey Tulyakov, and Mohamed Elhoseiny.
\newblock Stylegan-v: A continuous video generator with the price, image
  quality and perks of stylegan2.
\newblock In {\em CVPR}, 2022.

\bibitem{song2023latent}
Yue Song, Andy Keller, Nicu Sebe, and Max Welling.
\newblock Latent traversals in generative models as potential flows.
\newblock In {\em ICML}. PMLR, 2023.

\bibitem{song2022improving}
Yue Song, Nicu Sebe, and Wei Wang.
\newblock Improving covariance conditioning of the svd meta-layer by
  orthogonality.
\newblock In {\em ECCV}, 2022.

\bibitem{song2022orthogonal}
Yue Song, Nicu Sebe, and Wei Wang.
\newblock Orthogonal svd covariance conditioning and latent disentanglement.
\newblock {\em IEEE TPAMI}, 2022.

\bibitem{spingarn2021gan}
Nurit Spingarn-Eliezer, Ron Banner, and Tomer Michaeli.
\newblock Gan" steerability" without optimization.
\newblock {\em ICLR}, 2021.

\bibitem{tzelepis2021warpedganspace}
Christos Tzelepis, Georgios Tzimiropoulos, and Ioannis Patras.
\newblock Warpedganspace: Finding non-linear rbf paths in gan latent space.
\newblock In {\em ICCV}, 2021.

\bibitem{voynov2020unsupervised}
Andrey Voynov and Artem Babenko.
\newblock Unsupervised discovery of interpretable directions in the gan latent
  space.
\newblock In {\em ICML}, 2020.

\bibitem{wei2021orthogonal}
Yuxiang Wei, Yupeng Shi, Xiao Liu, Zhilong Ji, Yuan Gao, Zhongqin Wu, and
  Wangmeng Zuo.
\newblock Orthogonal jacobian regularization for unsupervised disentanglement
  in image generation.
\newblock In {\em ICCV}, 2021.

\bibitem{wu2021stylespace}
Zongze Wu, Dani Lischinski, and Eli Shechtman.
\newblock Stylespace analysis: Disentangled controls for stylegan image
  generation.
\newblock In {\em CVPR}, 2021.

\bibitem{xu2021linear}
Jianjin Xu and Changxi Zheng.
\newblock Linear semantics in generative adversarial networks.
\newblock In {\em CVPR}, 2021.

\bibitem{xu2022transeditor}
Yanbo Xu, Yueqin Yin, Liming Jiang, Qianyi Wu, Chengyao Zheng, Chen~Change Loy,
  Bo Dai, and Wayne Wu.
\newblock Transeditor: Transformer-based dual-space gan for highly controllable
  facial editing.
\newblock In {\em CVPR}, 2022.

\bibitem{yang2021semantic}
Ceyuan Yang, Yujun Shen, and Bolei Zhou.
\newblock Semantic hierarchy emerges in deep generative representations for
  scene synthesis.
\newblock {\em IJCV}, 2021.

\bibitem{yang2021discovering}
Huiting Yang, Liangyu Chai, Qiang Wen, Shuang Zhao, Zixun Sun, and Shengfeng
  He.
\newblock Discovering interpretable latent space directions of gans beyond
  binary attributes.
\newblock In {\em CVPR}, 2021.

\bibitem{yu2015lsun}
Fisher Yu, Ari Seff, Yinda Zhang, Shuran Song, Thomas Funkhouser, and Jianxiong
  Xiao.
\newblock Lsun: Construction of a large-scale image dataset using deep learning
  with humans in the loop.
\newblock {\em arXiv preprint arXiv:1506.03365}, 2015.

\bibitem{zhang2018stabilizing}
Jiong Zhang, Qi Lei, and Inderjit Dhillon.
\newblock Stabilizing gradients for deep neural networks via efficient svd
  parameterization.
\newblock In {\em ICML}, 2018.

\bibitem{zhang2018unreasonable}
Richard Zhang, Phillip Isola, Alexei~A Efros, Eli Shechtman, and Oliver Wang.
\newblock The unreasonable effectiveness of deep features as a perceptual
  metric.
\newblock In {\em CVPR}, 2018.

\bibitem{zhang2017s3fd}
Shifeng Zhang, Xiangyu Zhu, Zhen Lei, Hailin Shi, Xiaobo Wang, and Stan~Z Li.
\newblock S3fd: Single shot scale-invariant face detector.
\newblock In {\em ICCV}, 2017.

\bibitem{zhu2021low}
Jiapeng Zhu, Ruili Feng, Yujun Shen, Deli Zhao, Zheng-Jun Zha, Jingren Zhou,
  and Qifeng Chen.
\newblock Low-rank subspaces in gans.
\newblock {\em NeurIPS}, 2021.

\bibitem{zhu2022region}
Jiapeng Zhu, Yujun Shen, Yinghao Xu, Deli Zhao, and Qifeng Chen.
\newblock Region-based semantic factorization in gans.
\newblock {\em ICML}, 2022.

\bibitem{zhu2020learning}
Xinqi Zhu, Chang Xu, and Dacheng Tao.
\newblock Learning disentangled representations with latent variation
  predictability.
\newblock In {\em ECCV}. Springer, 2020.

\end{thebibliography}
}

\appendix

\section{Limitation and Future Work}

Our current experiments only validate our approach in fine-tuning StyleGANs. Despite the easy usage, the image fidelity and disentanglement performance might be better if we could train StyleGANs equipped with our Householder Projector from scratch. However, due to the limited computational resources, this point cannot be validated for now. Additionally, in the current setting, we pre-define the number of semantics of each layer to a fixed number (the rank of the projector). Seeking an adaptive scheme to automatically mine the semantics would be also an important direction of our future work. 

\section{Mathematical Derivation}

\subsection{Decomposing $\mathbf{U}$ and $\mathbf{V}$}

For $n{\times}n$ orthogonal matrix $\mathbf{U}$, there exists $\mathbf{H}_1\mathbf{H}_2\dots\mathbf{H}_n{=}\mathbf{U}$ where $\mathbf{H}_i$ is a Householder reflection matrix. The decomposition is achieved by the \textit{$n$-reflections theorem}: each $\mathbf{H}_i$ can be designed to zero out the non-diagonal entries of $\mathbf{U}$ in the $i$-th column and row and to set the diagonal entry to $1$. Such accumulation of $n$ reflectors can transform $\mathbf{U}$ into an identity matrix ($\mathbf{U}\mathbf{H}_n\dots\mathbf{H}_2\mathbf{H}_1{=} \mathbf{I}$). Since $\mathbf{H}_i$ is a reflection ($\mathbf{H}_i\mathbf{H}_i{=}\mathbf{I}$), this theorem directly gives the relation $\mathbf{U}{=}\mathbf{H}_1\mathbf{H}_2\dots\mathbf{H}_n$. 

\subsection{Orthogonality Preservation}

The orthogonality of a Householder matrix $\mathbf{H}_{i}$ can be easily verified by:
\begin{equation}
\begin{aligned}
    \mathbf{H}_{i}\mathbf{H}_{i}^{T} &=\Big(\mathbf{I}{-}2 \frac{\mathbf{h}_{i}\mathbf{h}_{i}^{T}}{||\mathbf{h}_{i}||_{2}^{2}}\Big) \Big(\mathbf{I}{-}2 \frac{\mathbf{h}_{i}\mathbf{h}_{i}^{T}}{||\mathbf{h}_{i}||_{2}^{2}}\Big)^{T}\\
    &=\frac{1}{||\mathbf{h}_{i}||_{2}^{4}} (||\mathbf{h}_{i}||_{2}^{2} \mathbf{I} - 2\mathbf{h}_{i}\mathbf{h}_{i}^{T})(||\mathbf{h}_{i}||_{2}^{2} \mathbf{I} - 2\mathbf{h}_{i}\mathbf{h}_{i}^{T})^{T} \\
    &=\frac{1}{||\mathbf{h}_{i}||_{2}^{4}} \Big(||\mathbf{h}_{i}||_{2}^{4}\mathbf{I} -4||\mathbf{h}_{i}||_{2}^{2}\mathbf{h}_{i}\mathbf{h}_{i}^{T}  + 4\mathbf{h}_{i}\mathbf{h}_{i}^{T}\mathbf{h}_{i}\mathbf{h}_{i}^{T} \Big)\\
    &=\frac{1}{||\mathbf{h}_{i}||_{2}^{4}} \Big(||\mathbf{h}_{i}||_{2}^{4}\mathbf{I} -4||\mathbf{h}_{i}||_{2}^{2}\mathbf{h}_{i}\mathbf{h}_{i}^{T}  + 4||\mathbf{h}_{i}||_{2}^{2}\mathbf{h}_{i}\mathbf{h}_{i}^{T} \Big) \\
    &= \frac{1}{||\mathbf{h}_{i}||_{2}^{4}} ||\mathbf{h}_{i}||_{2}^{4}\mathbf{I}  \\
    &= \mathbf{I}
\end{aligned}
\end{equation}
Similarly, when a gradient descent step is performed (\emph{i.e.,} $(\mathbf{h}_{i}{-}\eta\triangledown\mathbf{h}_{i}$), we still have the relation:
\begin{equation}
\begin{gathered}(\mathbf{I}{-}\frac{(\mathbf{h}_{i}{-}\eta\triangledown\mathbf{h}_{i})(\mathbf{h}_{i}{-}\eta\triangledown\mathbf{h}_{i})^{T}}{||\mathbf{h}_{i}{-}\eta\triangledown\mathbf{h}_{i}||_{2}^{2}})(\mathbf{I}{-}\frac{(\mathbf{h}_{i}{-}\eta\triangledown\mathbf{h}_{i})(\mathbf{h}_{i}{-}\eta\triangledown\mathbf{h}_{i})^{T}}{||\mathbf{h}_{i}{-}\eta\triangledown\mathbf{h}_{i}||_{2}^{2}})^{T}\\
    = \frac{1}{||\mathbf{h}_{i}{-}\eta\triangledown\mathbf{h}_{i}||_{2}^{4}}\Big(||\mathbf{h}_{i}{-}\eta\triangledown\mathbf{h}_{i}||_{2}^{4}\mathbf{I} \Big) = \mathbf{I}
\end{gathered}
\end{equation}
The orthogonality is preserved during the back-propagation and weight update phase.

\subsection{Householder Representation}

With the previous results on orthogonality preservation of a Householder matrix, we can proceed to show how an orthogonal matrix can be represented by the accumulation of elementary Householder reflectors. Given a square orthogonal eigenvector matrix defined as:
\begin{equation}
    \mathbf{U}=\sum^{d}_{i=1} \lambda_{i} \mathbf{u}_{i}\mathbf{u}_{i}^{T}    
\end{equation}
where $\mathbf{u_{i}}$ denotes the eigenvector of $\mathbf{U}$, and $\lambda_{i}\subset\{-1,1\}$ is the eigenvalue. Let $\prod^{d}_{j=1}\mathbf{H}_{j}$ be the accumulation of Householder reflectors as:
\begin{equation}
    \prod^{d}_{j=1}\mathbf{H}_{j} =\prod^{d}_{j=1}(\mathbf{I}-2\frac{\mathbf{h}_{j}\mathbf{h}_{j}^{T}}{||\mathbf{h}_{j}||_{2}^{2}})
\end{equation}
The eigenvector property directly gives
\begin{equation}
   \prod^{d}_{j=1}\mathbf{H}_{j}\mathbf{u}_{i}=\prod^{d}_{j=1}(\mathbf{I}-2\frac{\mathbf{h}_{j}\mathbf{h}_{j}^{T}}{||\mathbf{h}_{j}||_{2}^{2}})\mathbf{u}_{i}
   \label{eq:householder_times}
\end{equation}
If we set $\mathbf{h}_{j}=\mathbf{u}_{i}$ for $i=j$, the orthogonality would naturally lead to
\begin{equation}
\begin{gathered}
    (\mathbf{I}-2\frac{\mathbf{h}_{j}\mathbf{h}_{j}^{T}}{||\mathbf{h}_{j}||_{2}^{2}})\mathbf{u}_{i} = \mathbf{u}_{i},\ i\neq j \\
    (\mathbf{I}-2\frac{\mathbf{h}_{j}\mathbf{h}_{j}^{T}}{||\mathbf{h}_{j}||_{2}^{2}})\mathbf{u}_{i} = \lambda_{i}\mathbf{u}_{i},\ i = j
\end{gathered}
\end{equation}
Eq.~\eqref{eq:householder_times} is further simplified as:
\begin{equation}
    \prod^{d}_{j=1}\mathbf{H}_{j}\mathbf{u}_{i} =\lambda_{i}\mathbf{u}_{i}= \mathbf{U}\mathbf{u}_{i}
\end{equation}
The above equation shows that the relation $\mathbf{U}=\prod^{d}_{j=1}\mathbf{H}_{j}$ holds. This indicates that any orthogonal matrices can be represented by a series of Householder accumulations. 
 
\subsection{Semi-orthogonality of Non-Square Matrices}

For the fluency of text flow, we do not differentiate the projector $\mathbf{A}$ from square or non-square matrices in the paper. Strictly speaking, non-square matrices with orthonomal rows or columns (depending on whether $\mathbf{A}$ is a flat matrix or tall matrix) should be called semi-orthogonal matrices more precisely. Here we give a special note for the strictness of math definitions, but this does not influence the core contribution of our method or any experimental results. 

\section{Details of Datasets and Metrics}

\subsection{Datasets}

\noindent\textbf{StyleGAN2 Datasets.} FFHQ~\cite{kazemi2014one} consists of $70,000$ high-quality face images that have considerable variations in identifies and have good coverage in common accessories. LSUN Church~\cite{yu2015lsun} has $126,227$ scenes images of outdoor churches, and LSUN Cat~\cite{yu2015lsun} is comprised of $1,657,266$ different cat images collected online. 

\noindent\textbf{StyleGAN3 Datasets.} MetFaces~\cite{karras2020training} contains $1,336$ high-quality human faces extracted from works of arts. AFHQv2~\cite{choi2020stargan} is a dataset consisting of $15,803$  animal faces from three different domains, including cat, dog, and wildlife. SHHQv1~\cite{fu2022styleganhuman} covers $40,000$ images of diverse full-body clothed humans in its current version. Notice that their pre-trained models use $230,000$ images for training but only a subset of the training set is released. We expect that using the complete set for training would further improve the FID score of our method on SHHQ.


\subsection{Metrics}

\noindent\textbf{Fréchet Inception Distance (FID)~\cite{heusel2017gans}.} FID assesses the Fréchet distance of deep features between the set of generated images and the set of real images. More formally, given the feature distribution $\mathcal{N}(\mu,\Sigma)$ of real images and the feature distribution $\mathcal{N}(\mu',\Sigma')$ of fake images, the distance is computed as:
\begin{equation}
    d_{F}=\sqrt{||\mu-\mu'||_{2}^{2}+{\rm tr}\Big(\Sigma+\Sigma'-2(\Sigma^{\frac{1}{2}}\Sigma'\Sigma^{\frac{1}{2}})^{\frac{1}{2}}\Big)}
\end{equation}
A small value would indicate that the distance between distributions is close and the generated images are realistic. Our FID score is computed based on $50,000$ samples.

\noindent\textbf{Perceptual Path Length (PPL)~\cite{karras2019style} and Perceptual Interpretable Path Length (PIPL).} PPL subdivides the interpolation path into linear segments and measures the perceptual image distance of the segmented path. Let $\mathbf{w}_{1}$ and $\mathbf{w}_{2}$ be the randomly sampled latent code in the $\mathcal{W}$ space of StyleGANs. Then PPL defined in the $\mathcal{W}$ space is calculated as:
\begin{equation}
\begin{split}
    {\rm PPL}_{\mathcal{W}}=\mathbb{E}\Big[ \frac{1}{\epsilon^2} d(G( {\rm lerp}(\mathbf{w}_{1},\mathbf{w}_{2},t),\\
    G({\rm lerp}(\mathbf{w}_{1},\mathbf{w}_{2},t+\epsilon)      )) \Big]
\end{split}
\end{equation}
where $d(\cdot)$ represents the LPIPS~\cite{zhang2018unreasonable} distance, ${\rm lerp}(\cdot)$ denotes the spherical interpolation function, $t$ is a random variable sampled from $U(0,1)$, and $\epsilon$ is the subdivision constant, respectively. The division coefficient $\epsilon$ is set to $1e-4$ for all the experiments. 

The metric PPL suits use cases where the latent code is randomly interpolated. However, when the latent code is moved around as $\mathbf{z}+\mathbf{n}$ where $\mathbf{n}\in\mathbb{R}^{d}$ is an interpretable direction sampled from a given vector set (\emph{i.e.,} the eigenvectors extracted by SeFa~\cite{shen2021closed}), the PPL score can not reflect the smoothness of latent space. To make the score adapt to such vector-based manipulations, we propose our PIPL metric by naturally incorporating orthogonal vector perturbations into PPL. Formally, the PIPL is defined as:
\begin{equation}
    \begin{split}
    {\rm PIPL}_{\mathcal{W}}=\mathbb{E}\Big[ \frac{1}{\epsilon^2} d(G( {\rm lerp}(\mathbf{w}_{1},\mathbf{w}_{2},t),\\
    G({\rm lerp}(\mathbf{w}_{1},\mathbf{w}_{2},t)+\epsilon\mathbf{n}      )) \Big]
\end{split}
\end{equation}
where $\mathbf{n}$ is an orthogonal vector (\textit{i.e.,} $\mathbf{n}^{T}\mathbf{n}=1$) sampled from the given vector set. Here different vector sets are used because each model is fine-tuned and the interpretable directions are changed. It is thus more reasonable to use the corresponding directions of each method for evaluation.
Since the impact of orthogonal vector perturbation is very small in the perceptual distance change, we set $\epsilon$ as $1$ for StyleGAN2 and as $1e{-}2$ for StyleGAN3 to avoid the magnification by $\nicefrac{1}{\epsilon^2}$. We use different $\epsilon$ for StyleGAN2 and StyleGAN3 because these two models have different levels of sensitivities to the latent perturbation. StyleGAN3 is less sensitive due to the intrinsic equivariance properties and also the fact that we insert fewer layers. Compared with PPL, our proposed PIPL can better assess the vector-based latent disentanglement approaches. Both PPL and PIPL are computed with $10,000$ samples.

\noindent\textbf{Face Attribute Correlation.} For the attribute correlation, we first use S3FD~\cite{zhang2017s3fd} to extract the face region and then compute the normalized Pearson’s correlation between the traversal steps and the predictions using several pre-trained attributes estimators, including FairFace~\cite{karkkainen2021fairface} for face attributes (age, race, glasses, and gender) and HopeNet~\cite{doosti2020hope} for face poses. Among the pool of interpretable directions, the direction with the highest correlation is deemed to control the attribute. The results are averaged based on $2K$ same samples generated by PTI~\cite{roich2022pivotal}.


\section{Ablation Studies}

This section presents the ablations on studying the impact of fine-tuning steps, batch size, initialization schemes, low-rank orthogonality, and acceleration techniques.


\begin{table}[tbp]
    \centering
    \resizebox{0.85\linewidth}{!}{
    \begin{tabular}{c|c|c|c}
    \toprule
         \textbf{Steps} & \textbf{FID} ($\downarrow$) & \textbf{PPL} ($\downarrow$) & \textbf{PIPL} ($\downarrow$) \\
         \midrule
         $0\%$ &18.97 &799.38 & 0.101\\
         $0.25\%$  &10.56 &427.90 &0.057 \\ 
         $0.5\%$  &9.31 &474.12 &0.060 \\ 
         $1\%$ & 8.46 & 526.26 & 0.057\\ 
         $2\%$ &8.10 &544.31 & 0.056 \\ 
         \midrule\midrule
         Original StyleGAN2 &8.37 &722.24 &0.141 \\
    \bottomrule
    \end{tabular}
    }
    \caption{Impact of different fine-tuning steps ($\%$ of the original training steps) on LSUN Cat~\cite{yu2015lsun} with StyleGAN2~\cite{karras2020analyzing}.}
    \label{tab:ablation_step}
\end{table}

\subsection{Impact of Fine-tuning Steps.} Table~\ref{tab:ablation_step} evaluates the impact of fine-tuning steps on the performance. When the number of fine-tuning steps increases, the FID score and the image fidelity improve. However, the PPL smoothness deteriorates as FID improves. This can be considered as a trade-off between image quality and latent smoothness. We choose $1\%$ fine-tuning steps to avoid incurring large computational burdens. Nonetheless, one can always choose an appropriate step if a better FID score is required.

\subsection{Impact of Batch Size}

\begin{table}[htbp]
    \centering
     \resizebox{0.89\linewidth}{!}{
    \begin{tabular}{c|c|c|c}
    \toprule
         \diagbox{\textbf{BS}}{\textbf{Metrics}} & \textbf{FID} ($\downarrow$) & \textbf{PPL} ($\downarrow$) & \textbf{PIPL} ($\downarrow$) \\
    \midrule
          8&5.78 &468.99  & 0.029\\
          16&4.94 &473.19 & 0.031 \\
          32&3.72 &457.52 & 0.030 \\
    \bottomrule
    \end{tabular}
    }
    \caption{Impact of Batch Size (BS) on the quality of generated images on LSUN Church~\cite{yu2015lsun} with StyleGAN2~\cite{karras2020analyzing}.}
    \label{tab:ablation_bs}
\end{table}

Table~\ref{tab:ablation_bs} presents the image fidelity and the latent space smoothness when different batch sizes are used for fine-tuning. When the batch size increases, the FID score has also steady improvements, while the latent space smoothness is mildly influenced. This indicates that the batch size can greatly affect image quality. We believe that using a larger batch size can further boost the FID score of our method, particularly in StyleGAN3 experiments where our batch size is actually smaller than the original setting due to computational resources.

\begin{table}
    \centering
    \resizebox{0.99\linewidth}{!}{
    \begin{tabular}{c|c|c|c}
    \toprule
         \textbf{Initialization Scheme}  & \textbf{FID} ($\downarrow$) & \textbf{PPL} ($\downarrow$) & \textbf{PIPL} ($\downarrow$) \\
    \midrule
         Random Initialization      &4.89 &978.79 &0.160 \\
         Nearest-orthogonal Mapping &\textbf{4.40}  &\textbf{966.23}  &\textbf{0.141} \\
    \bottomrule
    \end{tabular}
    }
    \caption{Impact of initialization schemes on FFHQ~\cite{kazemi2014one}.}
    \label{tab:ablation_intia}
\end{table}

\begin{table}
    \centering
     \resizebox{0.79\linewidth}{!}{
    \begin{tabular}{c|c|c}
    \toprule
         \textbf{\makecell[c]{Computation \\ Method}} & \textbf{\makecell[c]{Vanilla\\ Accumulation}} & \textbf{\makecell[c]{Accelerated\\ Accumulation}} \\
         \midrule
         Time (ms)  & 68.02 & \textbf{2.67}\\
    \bottomrule
    \end{tabular}
    }
    \caption{Computation time cost for Householder accumulation of representing $512{\times}512$ matrices measured on a RTX A6000 GPU.}
    \label{tab:acceleration}
\end{table}

\subsection{Impact of Initialization and Acceleration.} Table~\ref{tab:ablation_intia} compares the performance of different initialization schemes. The proposed nearest-orthogonal initialization maps the pre-trained projector into the nearest orthogonal form, which leverages the statistic of well-trained network weights. It thus outperforms the ordinary random initialization. Table~\ref{tab:acceleration} shows the computational time of our accelerated Householder aggregation. The acceleration technique significantly improves $25$ times the speed of vanilla accumulation, enabling efficient implementation of our Householder Projector in deep neural networks. The marginal time cost would not bring much computational overhead to generative models.


\subsection{Impact of Low-rank Orthogonality}

\begin{table}[htbp]
    \centering
     \resizebox{0.79\linewidth}{!}{
    \begin{tabular}{c|c|c|c}
    \toprule
         \textbf{Rank}  & \textbf{FID} ($\downarrow$) & \textbf{PPL} ($\downarrow$) & \textbf{PIPL} ($\downarrow$) \\
         \midrule
         512 (full rank) &4.34 &390.89 & 0.025 \\
         10 (low rank)& 3.72 &457.52 & 0.030\\
         5 (low rank) &3.65 &461.76 & 0.032 \\
    \bottomrule
    \end{tabular}
    }
     \caption{Impact of different matrix rank for our Householder Projector on LSUN Church~\cite{yu2015lsun} with StyleGAN2.}
    \label{tab:rank}
\end{table}

\begin{figure}
    \centering
    \includegraphics[width=.99\linewidth]{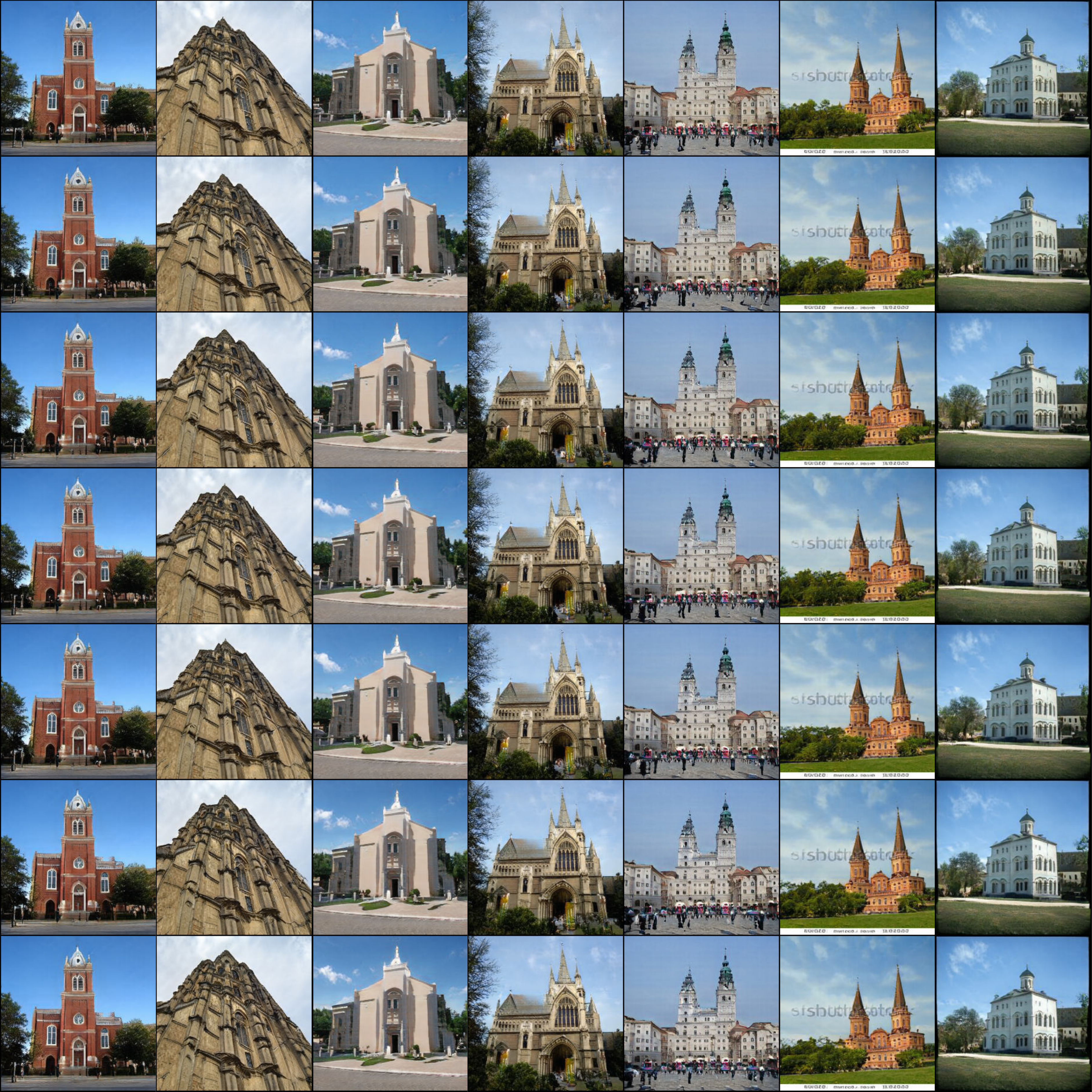}
    \caption{Exemplary latent traversal results of full-rank Householder Projector on LSUN Church~\cite{yu2015lsun} with StyleGAN2~\cite{karras2020analyzing}. Due to the large dimensionality, using the full-rank projector would split data variations among the eigenvectors. The output changes are thus imperceptible and it is unlike to inspect the concrete semantic attribute of each traversal direction.}
    \label{fig:fullrank}
\end{figure}

Table~\ref{tab:rank} presents the quantitative evaluation results on the impact of projector rank. The FID score of the full-rank projector falls behind that of the low-rank projector. This stems from the fact the full-rank projector might be slower to converge and harder to optimize within the very limited fine-tuning steps. In terms of latent smoothness, the full-rank projector seems to outperform the low-rank projector. However, as shown in Fig.~\ref{fig:fullrank}, there are not much variations in the traversal results and it is hard to inspect the specific semantic attributes of the identified directions. In this case, the advantage of full-rank projector on PPL and PIPL might come from less meaningful variations instead of the improved latent smoothness. Setting the matrix rank to $5$ and $10$ leads to very competitive performance. We set the rank to $10$ throughout the experiments because we empirically observed that each layer of StyleGANs has approximately $10$ semantic concepts. Nonetheless, the readers are encouraged to set different ranks for other datasets and architectures if more semantics are observed.


\section{More Visualizations and Discussions}

\subsection{Semantic Unambiguity}

\begin{figure*}
    \centering
    \includegraphics[width=0.99\linewidth]{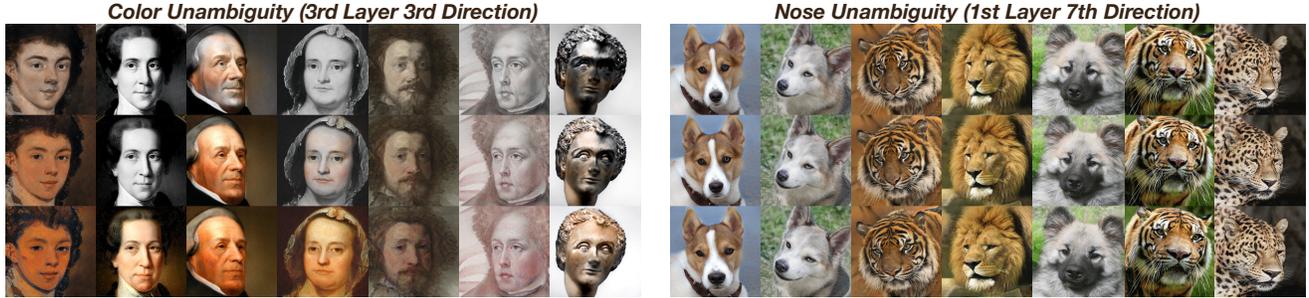}
    \caption{Illustration of semantic unambiguity on MetFaces~\cite{karras2020training} and AFHQ~\cite{choi2020stargan} based on StyleGAN3~\cite{karras2021alias} equipped with our Householder Projector. The discovered interpretable directions are semantically consistent among different samples.}
    \label{fig:seman_unambiguity}
\end{figure*}

Fig.~\ref{fig:seman_unambiguity} displays some examples of semantics unambiguity. The interpretable directions identified by our Householder Projector are unambiguous: different samples would have consistent semantic attribute changes when the latent code is moved by the discovered directions.   

\subsection{Semantic Hierarchy}


\begin{figure*}
    \centering
    \includegraphics[width=0.99\linewidth]{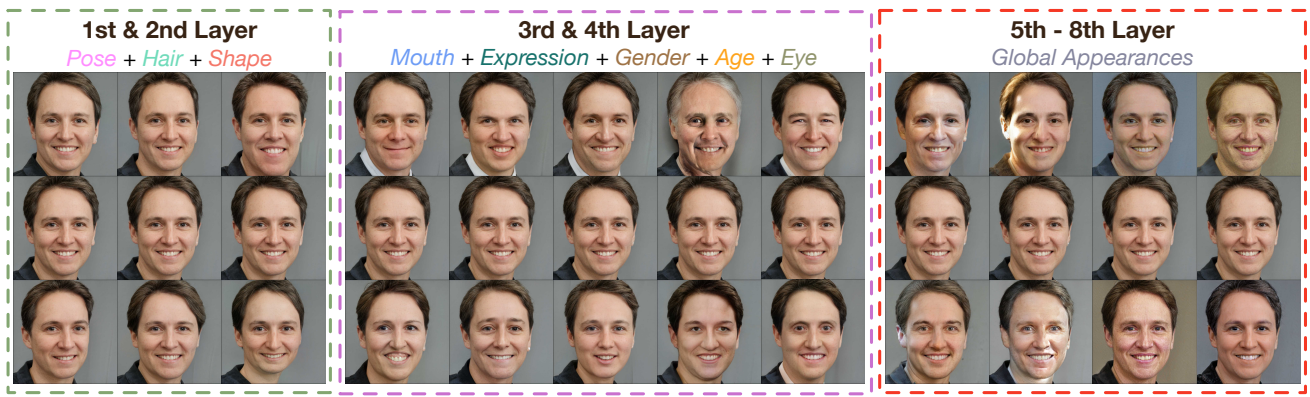}
    \caption{The layer hierarchy of semantic attributes identified by our Householder Projector based on FFHQ~\cite{kazemi2014one} with StyleGAN2~\cite{karras2020analyzing}.} 
    \label{fig:seman_hierachy}
\end{figure*}

Fig.~\ref{fig:seman_hierachy} shows the layer hierarchy of different semantics on FFHQ~\cite{kazemi2014one}. The shallow layers mainly focus on some geometric changes of the input images. Then the middle layers proceed to manipulate local details such as mouths and eyes. Finally, the deep layers target the global style and appearance of the images. Overall, the semantics hierarchy meets the same trend of StyleGANs. This indicates that our Householder Projector does not modify the semantics hierarchy of pre-trained models but tunes the model to mine more disentangled semantic concepts. 

\subsection{Semantic Diversity}

\begin{figure*}[htbp]
    \centering
    \includegraphics[width=0.99\linewidth]{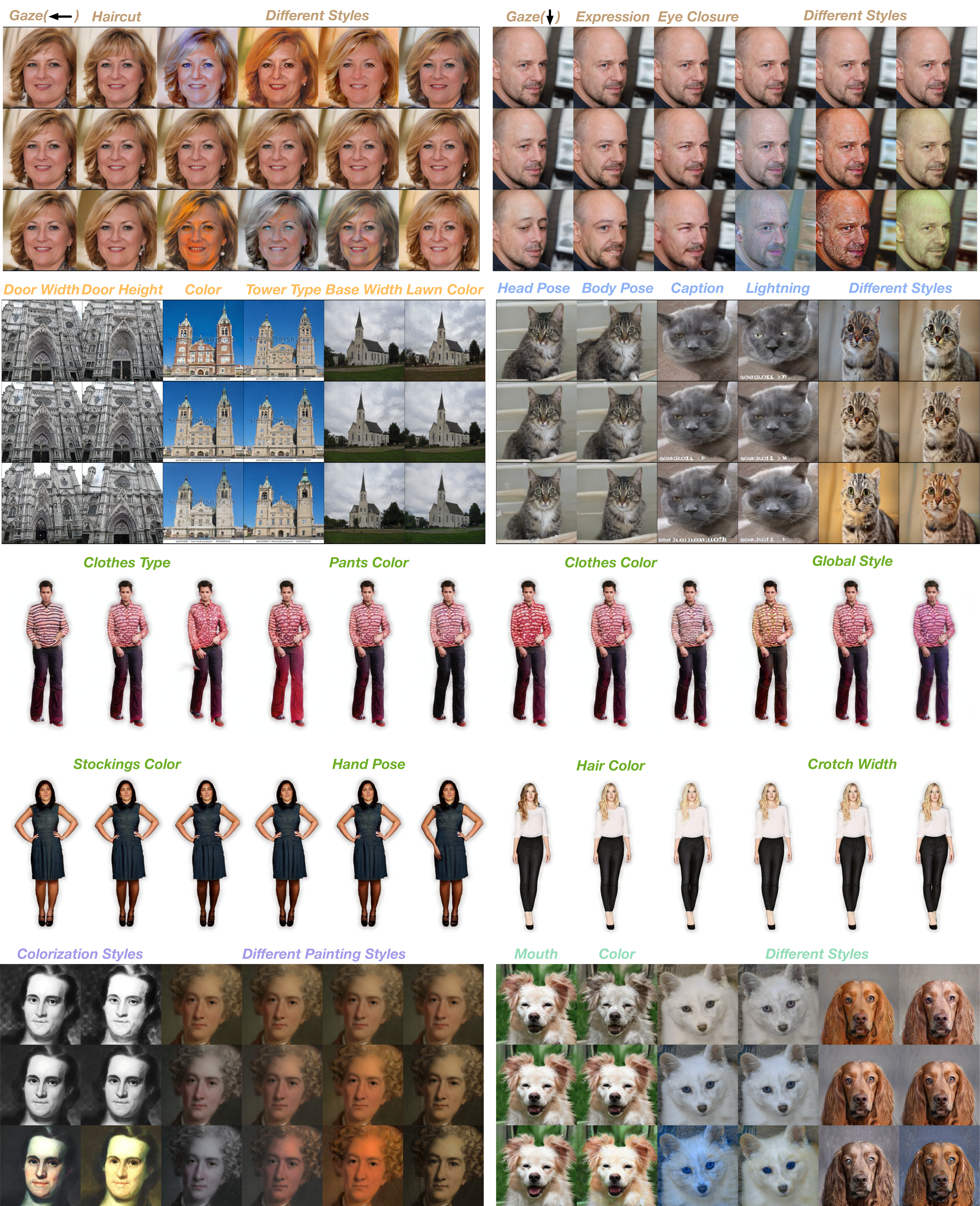}
    \caption{Gallery of more semantic attributes discovered on the used datasets. Here we display more style-related semantics.}
    \label{fig:diversity_app}
\end{figure*}

Fig.~\ref{fig:diversity_app} displays some more semantic attributes discovered on the used datasets. Different from the paper, here we exhibit more style semantics, \emph{i.e.,} the global appearance changes in the high-level layers of StyleGANs. Specific to each datatset, the style semantics correspond to different global variations that frequently occur in the datasets. For example, the style variations in MetFaces~\cite{karras2020training} are mainly different painting and colorization styles, and the style variations in FFHQ~\cite{kazemi2014one} mainly concern global color contrast, image sharpness, and different color temperatures. 

\subsection{Visual Comparison on Other Datasets}

\begin{figure*}
    \centering
    \includegraphics[width=0.99\linewidth]{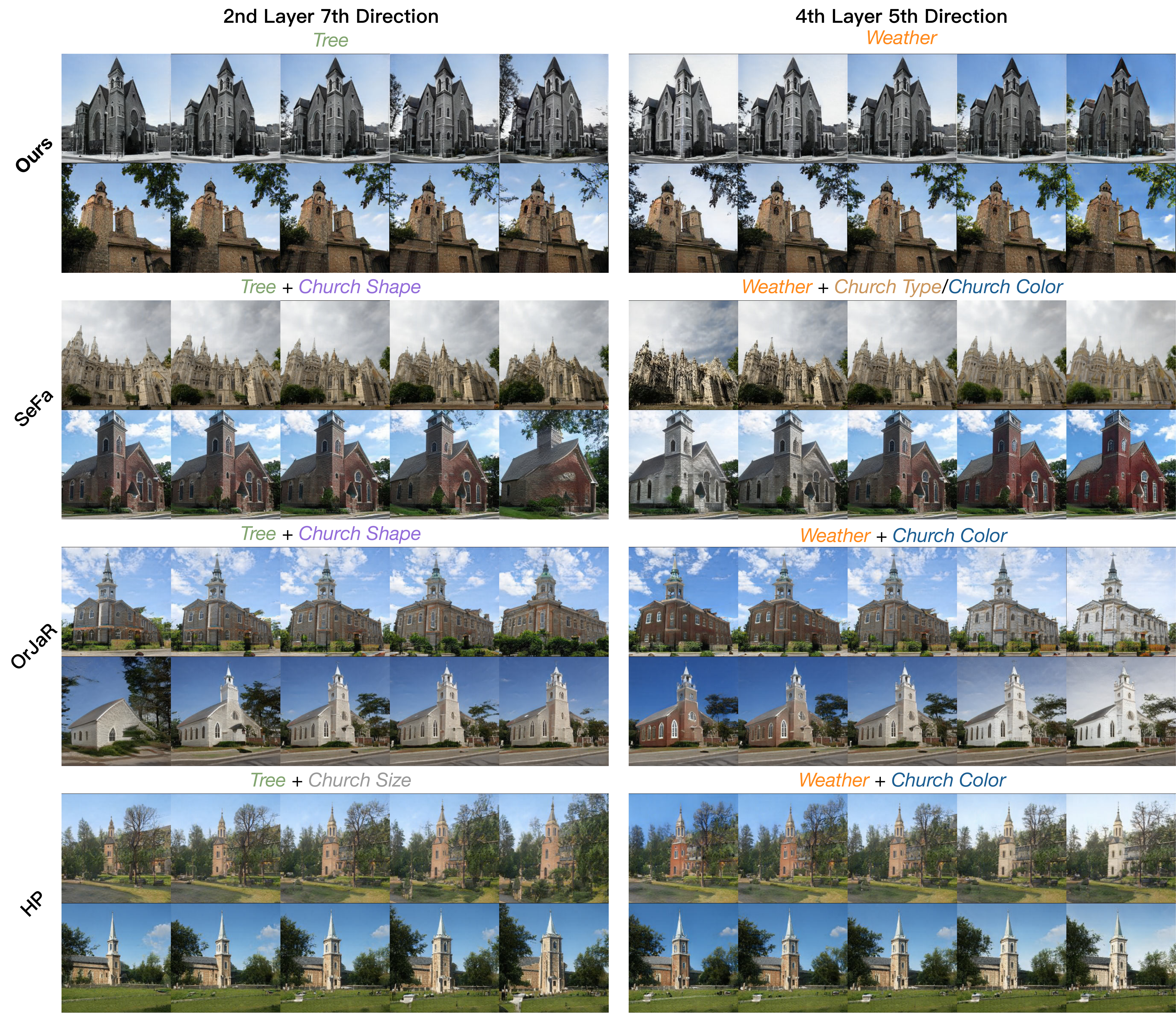}
    \caption{Exemplary latent traversal comparison of two attributes on LSUN Church~\cite{yu2015lsun}.}
    \label{fig:com_church}
\end{figure*}

\begin{figure*}
    \centering
    \includegraphics[width=0.99\linewidth]{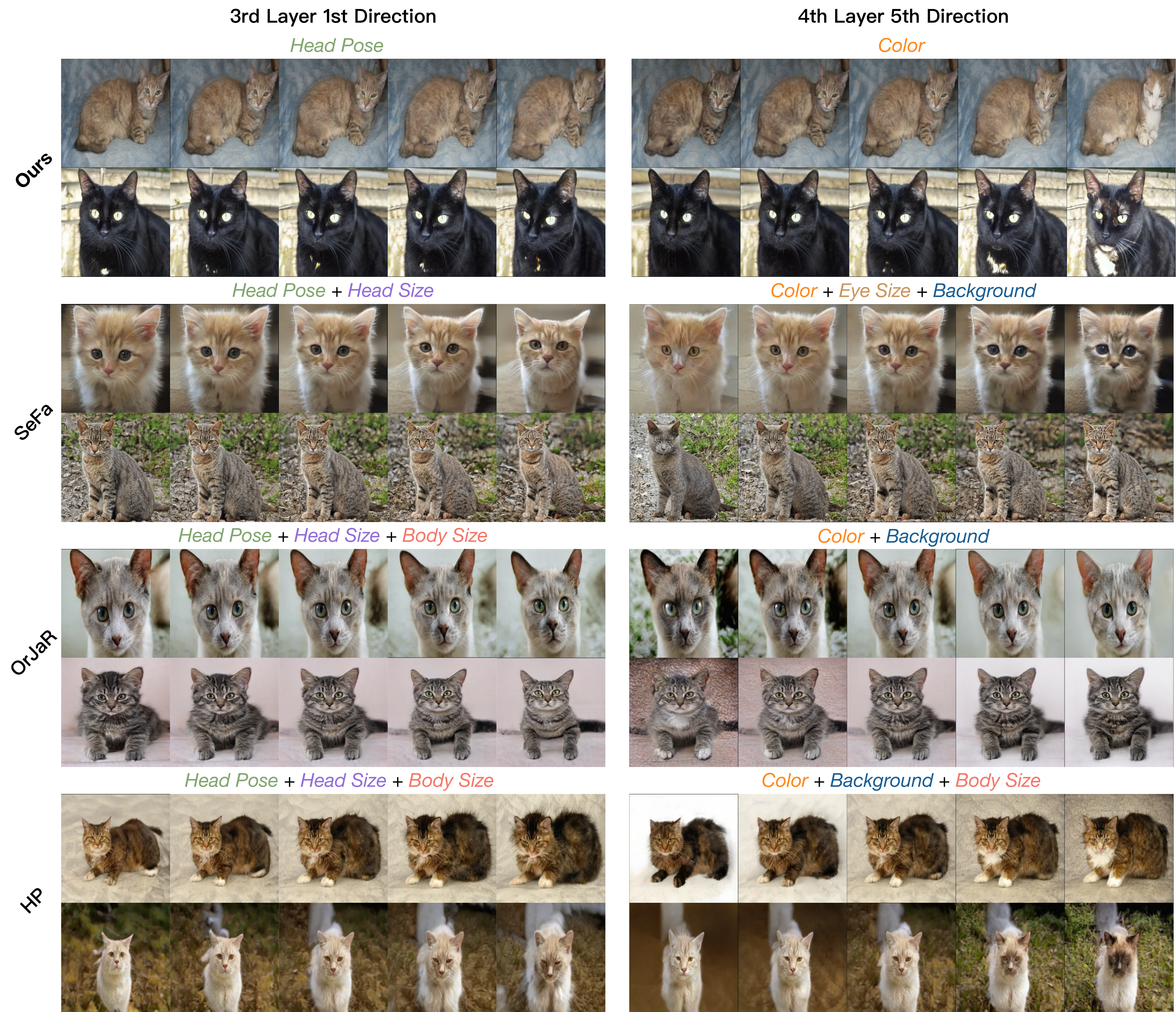}
    \caption{Exemplary latent traversal comparison of two attributes on LSUN Cat~\cite{yu2015lsun}.}
    \label{fig:com_cat}
\end{figure*}

\begin{figure*}
    \centering
    \includegraphics[width=0.99\linewidth]{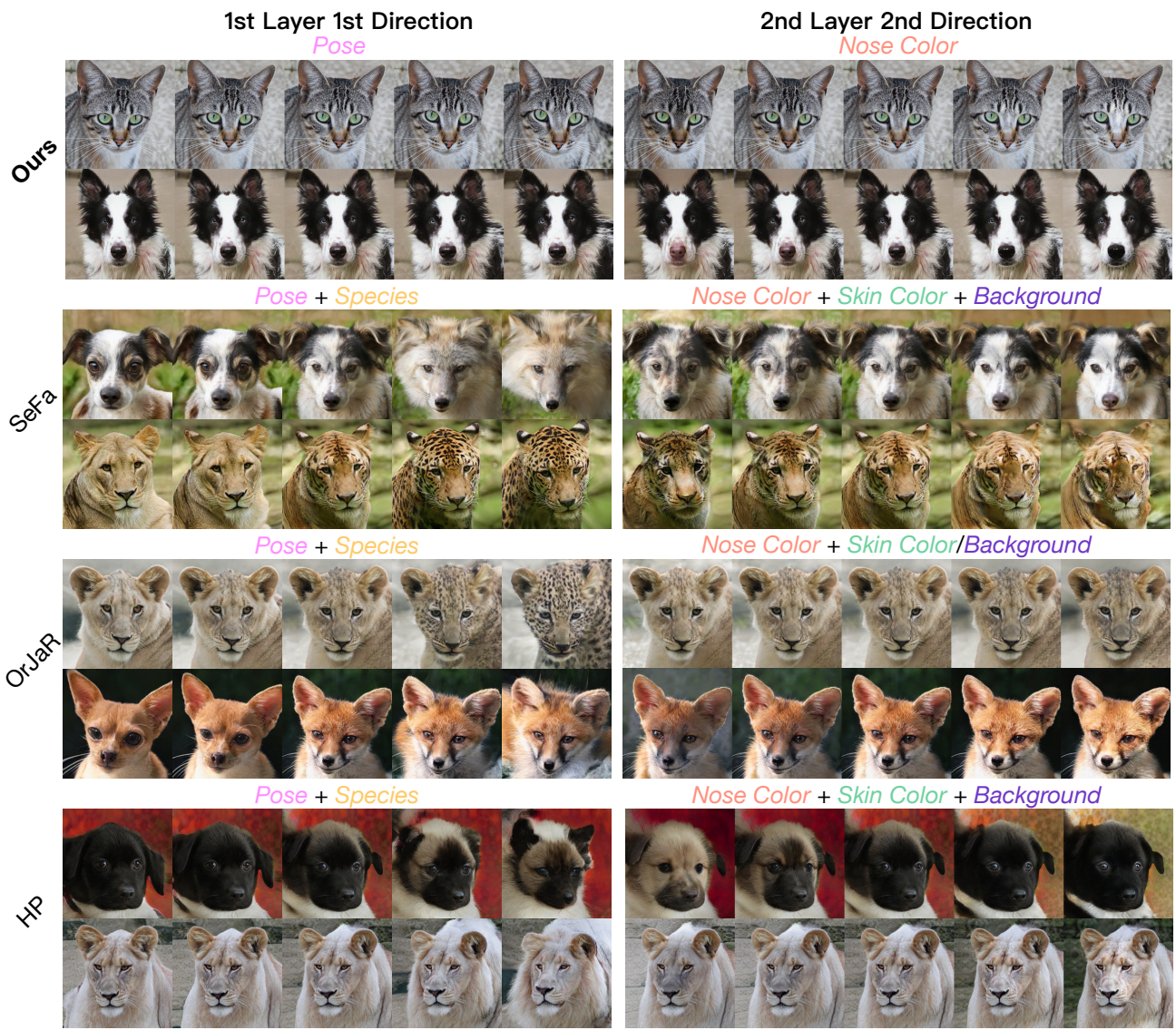}
    \caption{Exemplary latent traversal comparison of two attributes on AFHQv2~\cite{choi2020stargan}.}
    \label{fig:com_afhq}
\end{figure*}


\begin{figure*}
    \centering
    \includegraphics[width=0.99\linewidth]{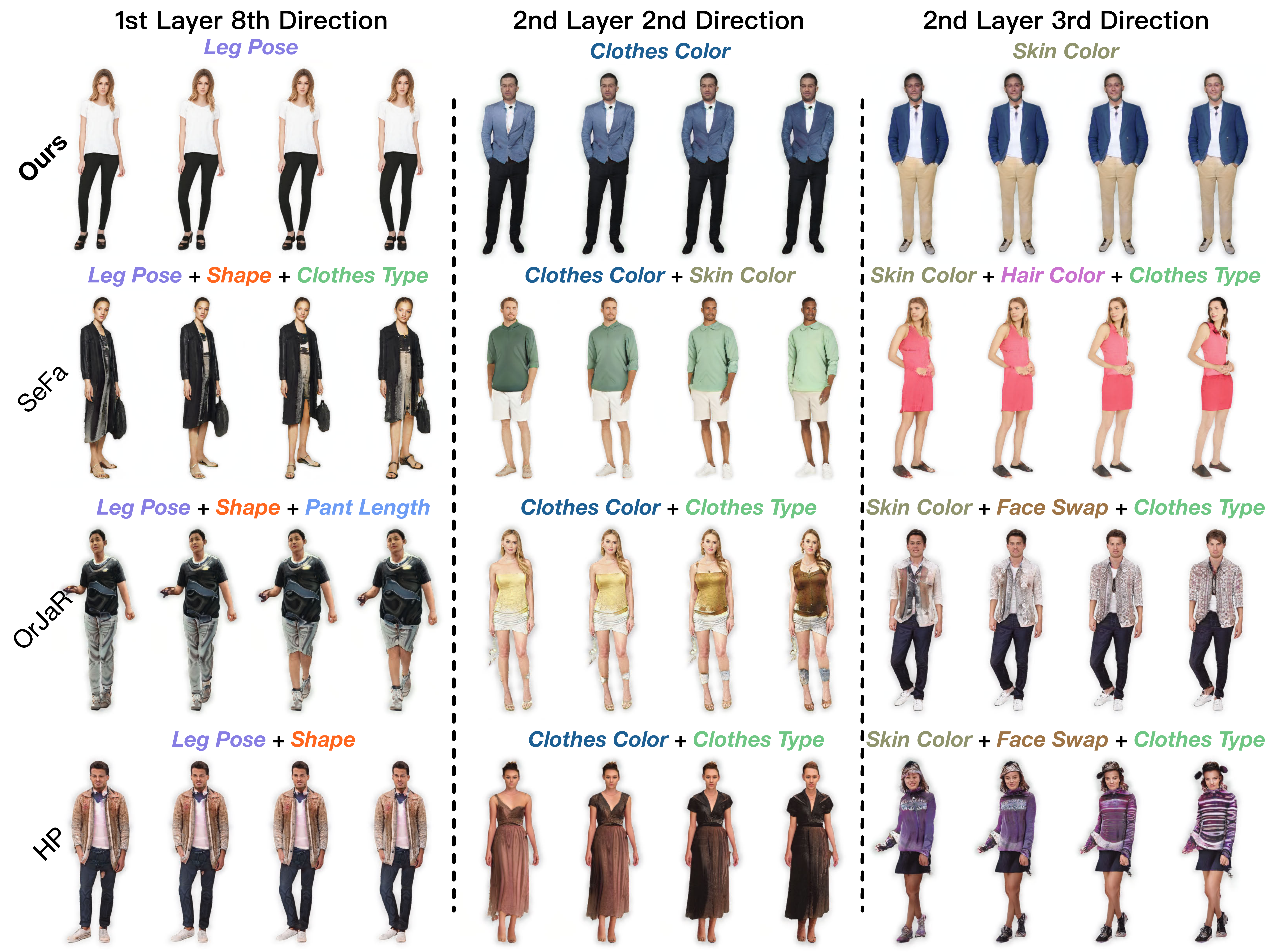}
    \caption{Exemplary latent traversal comparison of three attributes on SHHQ~\cite{fu2022styleganhuman}.}
    \label{fig:com_human_app}
\end{figure*}

\begin{figure*}
    \centering
    \includegraphics[width=0.99\linewidth]{imgs/com_ffhq_app.pdf}
    \caption{Exemplary latent traversal comparison of two attributes on FFHQ~\cite{kazemi2014one}.}
    \label{fig:com_ffhq_app}
\end{figure*}

Fig.~\ref{fig:com_church}, Fig.~\ref{fig:com_cat}, Fig.~\ref{fig:com_afhq}, Fig.~\ref{fig:com_human_app}, and Fig.~\ref{fig:com_ffhq_app} present the exemplary attribute comparison across all used datasets. The results are consistent with the visualizations in the paper. Our Householder Projector is able to identify more disentangled semantic attributes and gives users more precise control of the image attributes in the generation process.

\subsection{Comparison with EigenGAN}

\begin{figure*}
    \centering
    \includegraphics[width=0.99\linewidth]{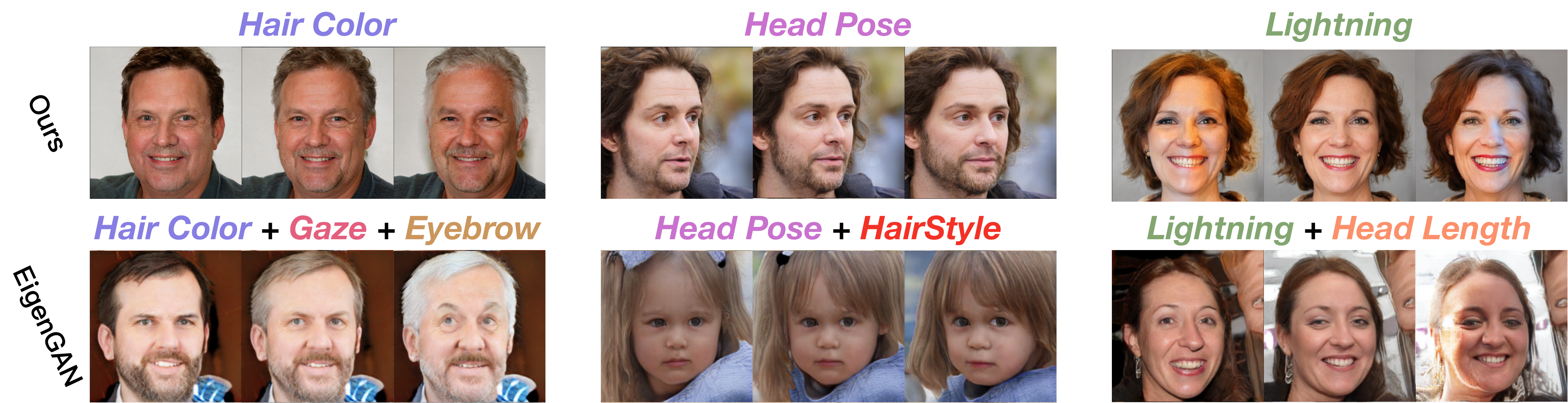}
    \caption{Comparison against EigenGAN~\cite{he2021eigengan} on some learned attributes with FFHQ~\cite{kazemi2014one}.}
    \label{fig:com_eigengan}
\end{figure*}

EigenGAN~\cite{he2021eigengan} is a small-scale GAN architecture that progressively injects orthogonal subspace into each layer of the generator to achieve disentanglement. Similar with HP~\cite{peebles2020hessian} and OrJaR~\cite{wei2021orthogonal}, the \emph{soft} orthogonality regularization is also used in EigenGAN to preserve the approximate orthogonality. Fig.~\ref{fig:com_eigengan} compares some semantics learned by our method and EigenGAN~\cite{he2021eigengan} on FFHQ~\cite{kazemi2014one}. Our method can discover more precise image attributes. 

\subsection{Generated Samples}

\begin{figure*}
    \centering
    \includegraphics[width=0.99\linewidth]{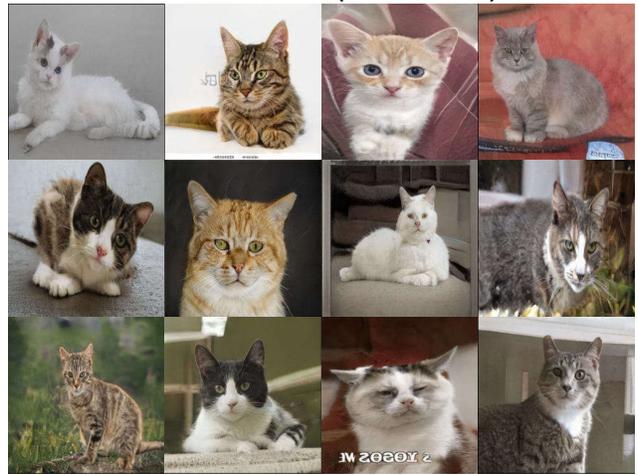}
    \caption{Random samples generated by StyleGANs~\cite{karras2020analyzing,karras2021alias} equipped with our Householder Projector. Our method does not harm the original quality of generate images.}
    \label{fig:samples}
\end{figure*}

Fig.~\ref{fig:samples} displays some samples randomly generated by our method across datasets. The image quality of the original StyleGANs~\cite{karras2020analyzing,karras2021alias} is maintained by our Householder Projector.

\end{document}